\documentclass[lettersize,journal]{IEEEtran}
\usepackage{amsmath,amsfonts}
\usepackage{algorithm}
\usepackage{array}
\usepackage[caption=false,font=normalsize,labelfont=sf,textfont=sf]{subfig}
\usepackage{textcomp}
\usepackage{stfloats}
\usepackage{url}
\usepackage{verbatim}
\usepackage{graphicx}
\usepackage{cite}
\usepackage[usenames]{color}
\definecolor{DarkBlue}{rgb}{0.0,0.08,0.6}
\definecolor{DarkRed}{rgb}{0.6,0.00,0.08}
\definecolor{DarkGreen}{rgb}{0.0,0.6,0.08}
\usepackage{lscape}
\usepackage[table]{xcolor}
\usepackage{amssymb}
\usepackage{makecell}
\usepackage{multirow}
\usepackage{rotating}
 \usepackage{psfrag}
\usepackage{bbding}
\definecolor{LightBlue}{rgb}{0.74,0.84,0.93}
\definecolor{LightGreen}{rgb}{0.77,0.95,0.7}
\definecolor{DarkBlue}{rgb}{0,0.44,0.75}
\usepackage[pagebackref=true,breaklinks=true,linkcolor=red,anchorcolor=green, citecolor=blue,letterpaper=true,colorlinks,bookmarks=false]{hyperref}

\usepackage[capitalize]{cleveref}
\crefname{section}{Sec.}{Secs.}
\Crefname{section}{Section}{Sections}
\Crefname{table}{Table}{Tables}
\crefname{table}{Tab.}{Tabs.}

\def\eg{\emph{e.g.}}
\def\ie{\emph{i.e.}}

\def\etc{\emph{etc.}}

\usepackage{booktabs} 
\usepackage{enumitem}
\usepackage{epsfig}
\usepackage{amsmath}
\usepackage{amssymb}

\usepackage{enumitem}
\usepackage{multirow}
\usepackage{multicol}
\usepackage{hhline}
\usepackage{pifont}
\newcommand{\xmark}{\ding{55}}%

\usepackage{algpseudocode}

\ifCLASSINFOpdf

\else

\fi

\begin{document}

\title{Boosting Convolutional Neural Networks with Middle Spectrum Grouped Convolution}

\author{Zhuo~Su, Jiehua Zhang, Tianpeng Liu, Zhen Liu, Shuanghui Zhang, Matti Pietik\"{a}inen, Li~Liu
\thanks{Corresponding author: Li Liu (\url{http://lilyliliu.com/}), Email: li.liu@oulu.fi}
\thanks{Zhuo Su, Jiehua Zhang, and Matti Pietik\"{a}inen are with the Center for Machine Vision and Signal Analysis at the University of Oulu, Finland. Email: zhuo.su@oulu.fi}
\thanks{Tianpeng Liu, Zhen Liu, Shuanghui Zhang, and Li Liu are with the College of Electronic Science, National University of Defense Technology, China. Li Liu is also with Center for Machine Vision and Signal analysis at the University of Oulu, Finland}
\thanks{This work was partially supported by National Key Research and Development Program of China No. 2021YFB3100800, the Academy of Finland under grant 331883 
and the National Natural Science Foundation of China under Grant 61872379 and 62022091. The CSC IT Center for Science, Finland, is also acknowledged for computational resources.}
}

\markboth{Submitted to IEEE Transactions on xxx}
{Su \MakeLowercase{\textit{et al.}}: Looking at Your Neighbors: Enhancing Binary Neural Networks with Local Binary Convolution}

\maketitle 

\begin{abstract}
This paper proposes a novel module called middle spectrum grouped convolution (MSGC) for efficient deep convolutional neural networks (DCNNs) with the mechanism of grouped convolution. It explores the broad ``middle spectrum'' area between channel pruning and conventional grouped convolution. Compared with channel pruning, MSGC can retain most of the information from the input feature maps due to the group mechanism; compared with grouped convolution, MSGC benefits from the learnability, the core of channel pruning, for constructing its group topology, leading to better channel division. The middle spectrum area is unfolded along four dimensions: group-wise, layer-wise, sample-wise, and attention-wise, making it possible to reveal more powerful and interpretable structures. As a result, the proposed module acts as a booster that can reduce the computational cost of the host backbones for general image recognition with even improved predictive accuracy. For example, in the experiments on ImageNet dataset for image classification, MSGC can reduce the multiply-accumulates (MACs) of ResNet-18 and ResNet-50 by half but still increase the Top-1 accuracy by more than $1\%$. With $35\%$ reduction of MACs, MSGC can also increase the Top-1 accuracy of the MobileNetV2 backbone. Results on MS COCO dataset for object detection show similar observations. Our code and trained models are available at \url{https://github.com/hellozhuo/msgc}.
\end{abstract}


\section{Introduction}
\label{sec:intro}

DCNN has revolutionized the computer vision community in many applications, from preliminary tasks like salient object detection~\cite{wang2021sodsurvey} and edge detection~\cite{su2021pdc}, to semantically more sophisticated tasks like image classification~\cite{li2017improving}, object detection~\cite{liu2020objectdetection}, and human pose estimation~\cite{chen2021anatomy}. The increasing prediction accuracy is usually at the cost of considerable consumed energies, with large computational cost by deep models~\cite{tan2019efficientnet}. How to reduce the computational cost of DCNNs without sacrificing accuracy has been a pressing topic, especially in the era of edge computing, since deep models are moving to resource constrained devices like smart phones and IoTs. In recent years, numerous efforts have been made in the community to tackle this issue, such as network pruning~\cite{lee2023fastfilterpruning, zhang2023adaptivefilterpruning}, compact and lighter network design~\cite{howard2017mobilenets,sandler2018mobilenetv2}, network quantization~\cite{zhang2022dynamicthreshold,su2022svnet}, \etc

\begin{figure}[t!]
\centering
    \centering
    \includegraphics[width=\linewidth]{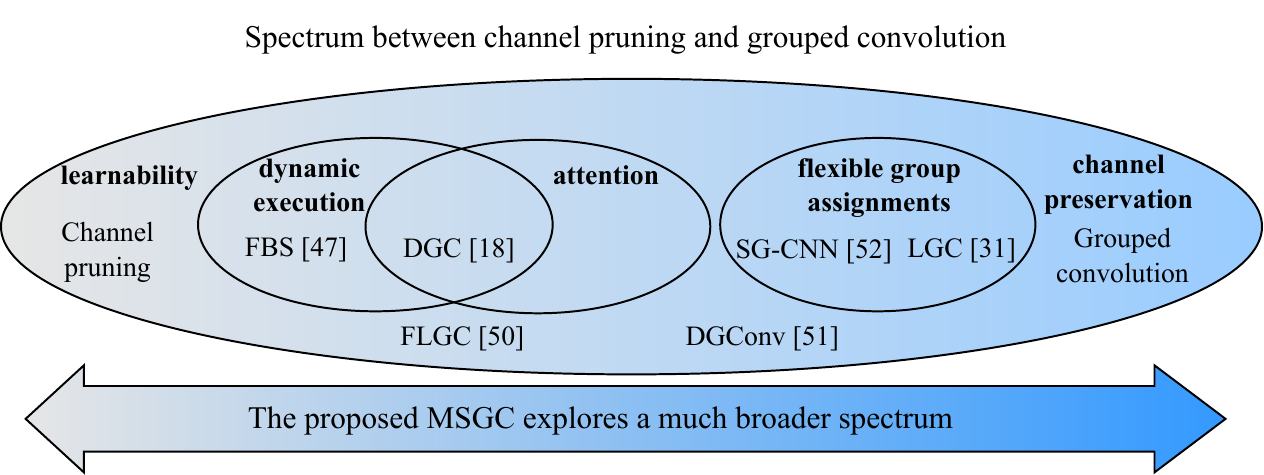}
    \caption{MSGC reveals a powerful network structure in the ``middle-spectrum'' area.}
    \label{fig:spectrum}
\end{figure}

Among these attempts, network pruning~\cite{lee2023fastfilterpruning,zhang2022whitebox,luo2017thinet,he2017channelpruing,han2016deepcompression} and grouped convolution~\cite{chen2021sharinggroup,su2020dgc,xie2017aggregated,ioannou2017deeproots,DBLP:igcv3} have attracted tremendous research interests. The former aims to prune the unnecessary redundant parts of deep models to make them lighter and more efficient to run, while the latter focuses on constructing compact structures by splitting computational operations into groups. It is not surprising that  many research works considered either of these two methods alone, since they are structurally independent. In this paper, we give a novel view by regarding them as two poles of a network designing spectrum (\cref{fig:spectrum}), inside which we find there are a big variety of structural possibilities that incorporate these two paradigms. Based on that, we further build our module that can effectively reveal powerful structures within the spectrum, outperforming both the previous channel pruning and grouped convolution based counterparts in both accuracy and computational cost. 
To make our motivation clearer, we start by giving a brief introduction on both methods below.

Without loss of generality, supposing a convolutional layer takes the input tensor $\mathbf{X}$ with $C$ channels and generates the output tensor $\mathbf{Y}$, the convolutional operation can be formulated as:

\begin{equation}
    \mathbf{Y} = [f^1(\pi^1(\mathbf{X})), f^2(\pi^2(\mathbf{X})), ..., f^G(\pi^G(\mathbf{X})],
\end{equation}
where $[,]$ represents the concatenation operation, $G$ is the number of groups, $f^i$ is a standard convolutional function that generates $1/G$ part of $\mathbf{Y}$, and $\pi^i$ is a selecting function extracting a sub-tensor from $\mathbf{X}$ with possibly fewer channels. 


When $G=1$ and $\pi\equiv \mathbf{1}\in {\{1\}}^{1\times C}$, which means all the input channels are kept as they are, the formulation reduces to the standard convolution. As the network often contains feature redundancy, it might be unnecessary to keep all the input channels. It is essentially the design spirit of many channel pruning methods, which focus on tuning $\pi$. 

The grouped convolution is formulated when we set $1<G \le C$ and the reduction of feature redundancy is neatly organized in a predefined way. For example, $\{\pi^i(\mathbf{X})|i=1,2,.., G\}$ are a series of regularly partitioned segments of the input tensor in the channel extent. This derived a lot of classical approaches in recent years like Deep Roots~\cite{ioannou2017deeproots}, ResNeXt~\cite{xie2017aggregated}, ShuffleNet~\cite{zhang2018shufflenet}, IGC series~\cite{zhang2017igcv1,xie2018igcv2,DBLP:igcv3}, UGConvs~\cite{zhao2019unitary}, ChannelNets~\cite{gao2020channelnets}, SCGNet~\cite{zhang2023scgnet}, sharing grouped convolution~\cite{chen2021sharinggroup}, and the extreme case of depthwise separable convolution used in the Inception modules~\cite{szegedy2015inceptionv1, szegedy2016inceptionv2}, MobileNet~\cite{howard2017mobilenets}, and Xception~\cite{chollet2017xception}, where $G=C$. One main focus of these methods is to design the correlation between groups to make the convolution complementary~\cite{DBLP:igcv3} (means that each output channel has at least one path to any of the input channels in the connection topology). Such encouragement of inter-group communication plays an important role in breaking the intrinsic order of channels~\cite{huang2018condensenet} to maximally preserve the prediction accuracy.

\begin{figure}[t!]
\centering
    \centering
    \includegraphics[width=\linewidth]{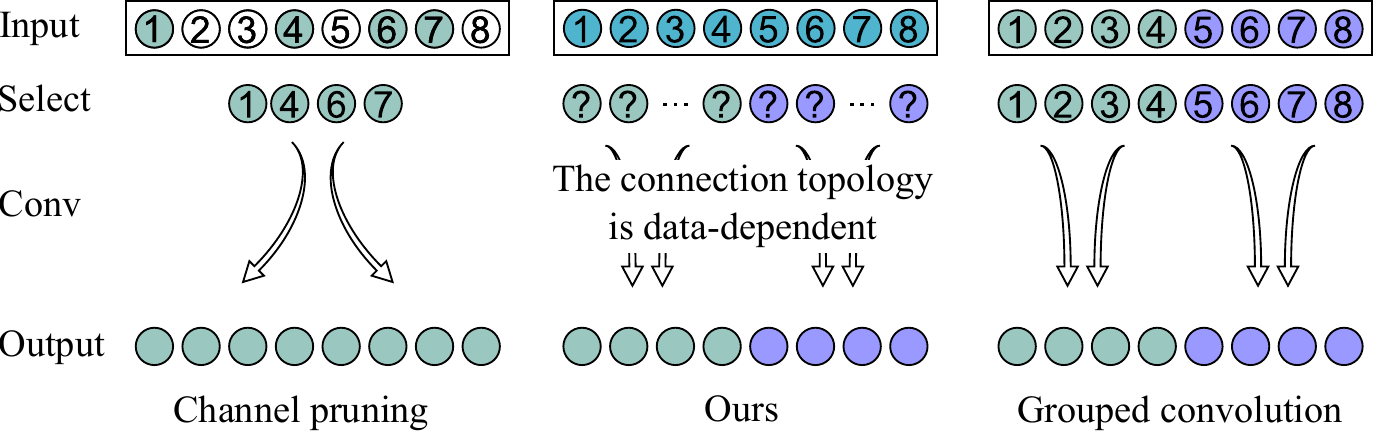}
    \caption{Illustration of the three paradigms.}
    \label{fig:msgc}
\end{figure}

However, we may rethink the relationship between the grouped convolution and channel pruning methods, since both aim to reduce the network redundancy from the channel extent. The design strategies behind them help us to derive the method proposed in the paper. The two frameworks are illustrated in~\cref{fig:msgc}.

On the one hand, channel pruning~\cite{zhang2022whitebox,zheng2022dncp,he2019fpgm,he2017channelpruing,luo2017thinet} attempts to learn the most important feature maps that contribute to the final prediction accuracy. Such \textbf{learnability} helps channel pruning to identify the unimportant channels that can be pruned without degrading the network performance significantly. However, since it is not easy to guarantee the correctness of such identification, or due to the bounded representation capacity by a pruning rate, there might always be some channels that contain specific useful and meaningful information than other channels. This is evidenced by the fact that a certain pruning rate usually causes a performance drop.

On the other hand, grouped convolution preserves all the input channels that may to some extent, avoid the above issue of information loss. In effect, different from channel pruning, grouped convolution hypothesizes that the grouped network can still learn enough information to give comparable prediction accuracy than the original network, by regularly sparsifying the connection but keeping all the input channels intact. In this way, the efforts on finding the most important channels, can be circumvented. In other words, grouped convolution is powered by \textbf{channel preservation}.

Generally, learning to prune and learning to group are manners that both lead to efficient and accurate networks. Previous works tend to consider them separately, which restrict their methods to go beyond.

We believe there is a better balance between information preservation and learnability that can be achieved by integrating both worlds toward building more powerful network structures. To achieve this, we proposed our module named MSGC (middle spectrum grouped convolution), which enables the learning process to find a structure in between, by unfolding the spectrum along four dimensions:

\begin{enumerate}
    \item Group-wise: injecting learnability for each group to learn how to segment channels;
    \item Layer-wise: allowing layer-dependent pruning ratios;
    \item Sample-wise: dynamically building grouped connection topology for individual samples (middle of \cref{fig:msgc});
    \item Attention-wise: decoupling channel gating and attention for individual groups.
\end{enumerate}

We conduct extensive experiments including image classification and object detection on large-scale datasets on the ResNet~\cite{he2016resnet}, MobileNetV2~\cite{sandler2018mobilenetv2}, and DenseNet~\cite{huang2017densenet} backbones, which consistently verify the superiority of MSGC compared with prior state-of-the-art methods. Those methods include both existing pruning-based and grouped convolution-based ones. Notably, we achieve not only the MAC reduction, but enhanced accuracy as well, due to the flexibility of forming groups. For example, on the ImageNet~\cite{imagenet} dataset for image classification, MSGC reduces computational cost of the ResNet backbones by 50\% and the MobileNetV2 backbone by 35\% with even improved accuracy. On the MS COCO~\cite{lin2014mscoco} dataset for object detection, MSGC can also effectively slim the backbones with negligible performance drop. In addition, MSGC can also be used to simply improve the prediction accuracy of the original networks with a relatively small pruning rate. In this case, the main role of MSGC transfers from an ``computation booster'' to a strong ``accuracy booster''.

The rest of this article is organized as follows. In \cref{sec:related_work}, we review the related works. Following that, we illustrate our method in detail in \cref{sec:method}. A comprehensive experimental comparison with state-of-the-art methods is provided in \cref{sec:experiments}, with detailed ablation studies. Finally, we conclude our paper in \cref{sec:conclusion}.

\section{Related work}
\label{sec:related_work}
\noindent \textbf{Network pruning} aims to prune the network weights or feature maps in convolutional layers to make the network less redundant, therefore less computational cost or memory storage is used to achieve the similar accuracy as before pruning. According to the pruning granularity, network pruning is generally categorized into structured pruning and unstructured pruning. The former prunes network by removing individual weights, leading to irregular structures after pruning~\cite{han2016deepcompression}. While the latter prunes groups of weights which belong to the whole filters, channels, or kernel blocks. Structured pruning keeps the pruned network in regular structures which can be more easily accelerated with existing deep learning frameworks on hardware~\cite{luo2017thinet}. Therefore, it obtains more research attention than unstructured pruning in recent years. Among different methods, the core is how to evaluate the importance of network weights, thereby to guide the pruning process. Many kinds of criteria were developed during last years, such as reconstruction error~\cite{he2017channelpruing,luo2017thinet}, $\ell_1$-norm and $\ell_2$-norm~\cite{DBLP:conf/iclr/l1norm,DBLP:sfp,zhang2023adaptivefilterpruning}, geometric median~\cite{he2019fpgm}, discrimination~\cite{DBLP:dcp,wang2019tmm-pruning}, channel sensitivity~\cite{ruan2021dpfps}, category-aware discrimination ability~\cite{zhang2022whitebox}, \emph{etc.} The way of pruning can also be searched~\cite{lee2023fastfilterpruning,zheng2022dncp,liu2019metapruning,ople2021tmm-pruning2} or learnt with a slimmable structure~\cite{DBLP:conf/iclr/slimmable,li2021dsnet}. In addition, for dynamic pruning, which our methods are most related to, the importance scores can be directly generated~\cite{DBLP:fbs,tang2021manidp}. 
The proposed MSGC belong to structured pruning, more specifically, channel pruning, where the pruned weights are the complete kernels connected to some certain channels. The importance scores of the network weights are data-dependent, which are calculated on-the-fly with dynamic execution.

\vspace{0.3em}
\noindent \textbf{Grouped convolution} dates back to AlexNet~\cite{NIPS2012_alexnet}, if not earlier, where the convolutional filters were put on two GPUs to suit the network's training memory. However, the by-product of this ``filter groups'' inspired a lot of following grouped convolution-based methods like \cite{ioannou2017deeproots,xie2017aggregated,zhang2018shufflenet}, to improve the performance and reduce the MACs of DCNNs. The main spirit of grouped convolution has been introduced in the previous section. 

\begin{figure*}[t!]
\centering
    \centering
    \includegraphics[width=\linewidth]{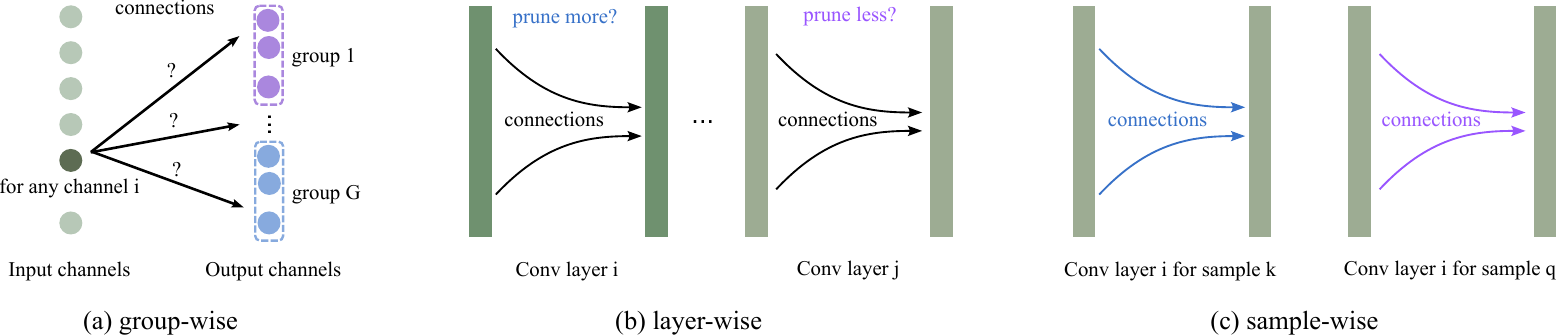}
    \caption{MSGC unfolds network structures along four dimensions during the learning process: (a) group-wise: input channels are assigned to certain group(s) in a learnable way; (b) layer-wise: the pruning rate automatically varies in different layers; (c) the connection topology varies with different samples even in the same layer. Please also see the illustration of attention-wise learning in our method part.}
    \label{fig:criteria}
\end{figure*}

\vspace{0.3em}
\noindent \textbf{MSGC-like methods}: Some previous works share certain similarities with our MSGC. Specifically, methods like LGC~\cite{huang2018condensenet}, FLGC~\cite{wang2019flgc}, and DGC~\cite{su2020dgc} can also preserve most or all of the input channels in a convolutional layer due to the group mechanism
and at the same time benefit from the learnability for channel partition. However, these methods suffer from their relatively narrow exploration scopes as shown in \cref{fig:spectrum}. For example, DGC~\cite{su2020dgc} generates dynamic connection topology but is constrained with a predefined pruning rate in each layer. FBS~\cite{DBLP:fbs} is a common dynamic channel pruning method without the consideration of grouped convolution. FLGC~\cite{wang2019flgc} and DGConv~\cite{zhang2019dgconv} actually shuffle the input channels (\ie, restrict each input channel belongs and only belongs to one group), and give fixed group topology. For LGC~\cite{huang2018condensenet} and SG-CNN~\cite{guo2020sgc}, they are only limited to the first dimension: learning how to divide the channels into groups.
In contrast, our method searches for a highly optimal structure from all four dimensions, which leads to better efficiency and predictive performance compared with both channel pruning- and grouped convolution-based methods. 

\vspace{0.3em}
\noindent \textbf{Dynamic network} or conditional network tunes its network topology or filters on the fly depending on the current input data, which can better increase its representation power and achieve a desired balance between accuracy and efficiency. The runtime execution of DCNNs can be implemented by dynamically skipping the layers~\cite{wu2018blockdrop,veit2018convnetaig}, re-weighting the filters~\cite{hu2018seblock,yang2019condconv}, pruning the channels or pixels~\cite{DBLP:fbs,dgnet,verelst2020dynamicconvlutionpixel}, decomposing the kernels~\cite{DBLP:conf/iclr/decomposing}, adjusting the nonlinear activations~\cite{chen2020dynamicrelu}, routing the inference paths~\cite{huang2018msdn}, cascading multiple networks~\cite{wang2020glance}. A more comprehensive review can be found in~\cite{han2021dynamicsurvey}. The proposed MSGC can be seen as a dynamic routing method that adjusts its connection topology among channels at runtime, a dynamic pruning method on the channel level, and a dynamic filter re-weighting method, in which we can use data-dependent attention for the selected channels.

\begin{figure*}[t!]
\centering
    \centering
    \includegraphics[width=\linewidth]{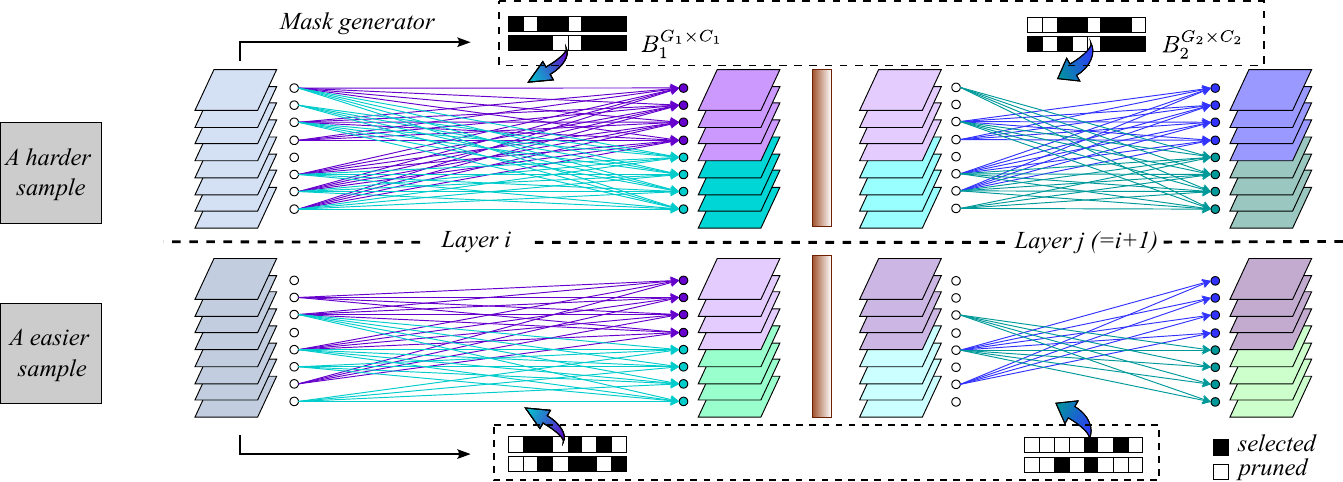}
    \caption{The exemplary illustration of a MSGC based network block with $M=2$ convolutional layers $L_1$ and $L_2$, where $G_1=G_2=2$, $C_1=C_2=8$. Here, the subscript ``$1$'' in $L_1$ or $G_1$ indicates the 1st layer in the block (which is actually ``Layer \emph{i}'' in the network). Please refer to the text for the meaning of the denotations. Best viewed in color.}
    \label{fig:overview}
\end{figure*}

\section{Methodology}
\label{sec:method}

\begin{figure*}[t!]
\centering
    \centering
    \includegraphics[width=\linewidth]{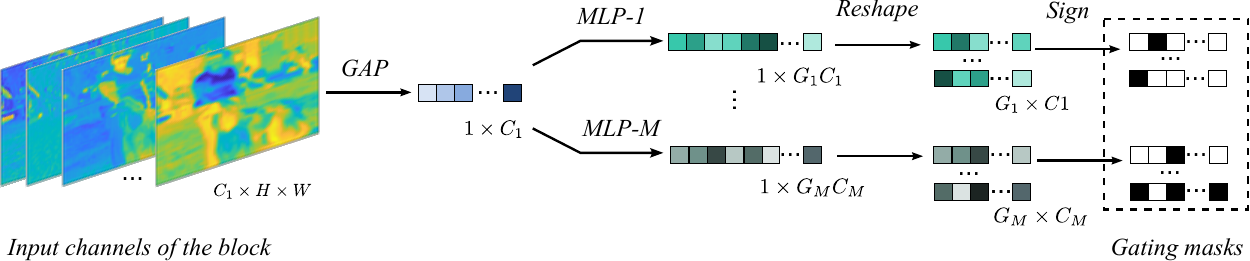}
    \caption{Structure of the mask generator in MSGC. Best viewed in color.}
    \label{fig:step1}
\end{figure*}


\subsection{Criteria}

To unfold the spectrum, namely, the possible structure candidates, from the four dimensions mentioned in \cref{sec:intro}, we aim to build powerful structures that conform to the following criteria:
\begin{enumerate}
    \item \textbf{Group-wise}: (\cref{fig:criteria} (a)) We inject learnability for each group to learn how to segment channels. It indicates a soft assignment of the input channels to each group. Therefore a more flexible group topology is learned during network training. It should be noted that our algorithm does not restrict each input channel to be assigned to only one group, but possibly zero, or multiple groups as well.
    
    \item \textbf{Layer-wise}: (\cref{fig:criteria} (b)) Different layers are enabled to have varying pruning ratios, due to the fact the each layer has its own property. For example, shallower and deeper layers may extract image abstraction in different semantic levels. Allowing layer-wise pruning ratios is expected to give better performance.
    
    \item \textbf{Sample-wise}: (\cref{fig:criteria} (c)) The connection topology (\eg, how the input and output channels are connected with each other) in groups would be dynamically determined depending on the individual input samples. It helps improve the adaptiveness of our network, \ie, enabling the network adapt to the varying characteristics of individual samples. The dynamic execution can also lead to a better optimization of resource allocation, as MSGC automatically learns to pay more computation for harder samples and vice versa, but keep the average computation at a low level.
    
    \item \textbf{Attention-wise}: In order to prune or select particular channels (\emph{i.e.}, gating), one typical way is to firstly generate a saliency vector/matrix with each element corresponding to a channel, then use the \emph{Top-K}~\cite{su2020dgc,DBLP:fbs} function or thresholds~\cite{DBLP:dst,wang2019flgc} to select the top ranking channels. After gating, the saliency values can be reused as the attention of the associated survived channels\footnote{By element-wise multiplying the feature map with the saliency value.}~\cite{su2020dgc,DBLP:fbs}. In this way, it essentially couples the functionality of gating and attention via the same saliency values. Instead, we decouple them in MSGC as two separate sub-modules to diversify the gating process.
    
\end{enumerate}



\subsection{Detailed design of MSGC}
\label{sec:pipeline}
The overview of the MSGC, which can be plugged into existing general convolutional backbones, is shown in \cref{fig:overview}.
Assuming we have a building block (\emph{e.g.}, a Bottleneck block in ResNet, a Reverted Bottleneck block in MobileNetV2, \etc) consisting of $M$ consecutive convolutional layers $\{L_i|i=1,2,...,M\}$, where the input and output tensor of $L_i$ have $C_i$ and $C_{i+1}$ channels, respectively (it should be noted that $L_1$ is denoted as the first layer of the block, not the network), and supposing the group number for each layer is $\{G_i|i=1,2,...,M\}$, the pipeline of building MSGC is detailed as follows:

\vspace{0.3em}
\noindent \textbf{Step 1: Generating binary masks.} To segment the input channels into groups for each layer in a learnable way, we use binary gating masks $\{\mathbf{B}_i^{G_i\times C_i}|i = 1, 2, ..., M\}$, where each row in $\mathbf{B}_i^{G_i\times C_i}$ only has 0/1 values representing which channels are selected (1) or not selected (0) to a certain group in layer $i$. For example, $\mathbf{B}_3^{2,5}=0$ means for the 3rd layer $L_3$, the 5th input channel is not assigned to the 2nd group. We build a mask generator to create the gating masks, where we simply take the input of the block (which is also the input of $L_1$) as the input of the generator. Since each channel is a $H\times W$ feature map, we use global average pooling (GAP) to firstly downsample the input to $\mathbb{R}^{1\times C_1}$, followed by $M$ light Multi-layer Perceptrons (MLPs) to generate $M$ saliency tensors $\{\mathbf{S}_i^{G_i\times C_i}|i = 1, 2, ..., M\}$ for each later layer respectively (including $L_1$), where $\mathbf{S}_i^{G_i\times C_i}\in\mathbb{R}^{G_i\times C_i}$.

Particularly, inspired by~\cite{hu2018seblock}, each MLP consists of two mapping matrices $W_1\in \mathbb{R}^{C_1 \times \frac{C_1}{R}}$ and $W_2 \in \mathbb{R}^{\frac{C_1}{R} \times G_i C_i}$, where $R$ is the reduction rate. The output of the MLP is then reshaped to $\mathbb{R}^{G_i \times C_i}$. We also insert a BN~\cite{ioffe2015bn} and ReLU layer between the two mappings to make the MLP a nonlinear transformation. 

Then, the Sign function ($\text{Sign}(x)=1$ if $x \geq 0$ and $=0$ otherwise) is used to convert each $\mathbf{S}_i^{G_i\times C_i}$ to $\mathbf{B}_i^{G_i\times C_i}$. The whole process is depicted in \cref{fig:step1}.

\vspace{0.3em}
\noindent \textbf{Step 2: Matching sub-filters and sub-input tensors for each group to conduct convolutions.} Specifically, the complete filters for a group in layer $L_i$ are $W\in \mathbb{R}^{k\times k \times C_i \times \frac{C_{i+1}}{G_i}}$, where $k$ is the kernel size. Based on the gating mask $\mathbf{B}_i^{G_i\times C_i}$, we take the sub filters along the ``$C_i$'' dimension, followed by the convolution with the correspondingly selected sub-input tensor. For example, supposing $L_i$ has 4 input and 8 output channels respectively and there are two groups in this layer, the 2nd row of $\mathbf{B}_i^{G_i\times C_i}$ is $[1, 0, 0, 1]$, we then select the 1st and 4th channel as the input for group 2. Meanwhile, the 5-8th output channels also belong to group 2 (output channels are regularly divided in groups). Thereby, the sub-filters that connect the 1st and 4th input channels to the 5-8th output channels are selected to conduct the convolution for group 2. Since the connection topology varies with different samples, all the filters will be kept in the memory to be selected and learnt with gradient descent. Therefore, we are able to preserve the whole capacity of the original network.

\begin{figure}[t!]
\centering
    \centering
    \includegraphics[width=0.8\linewidth]{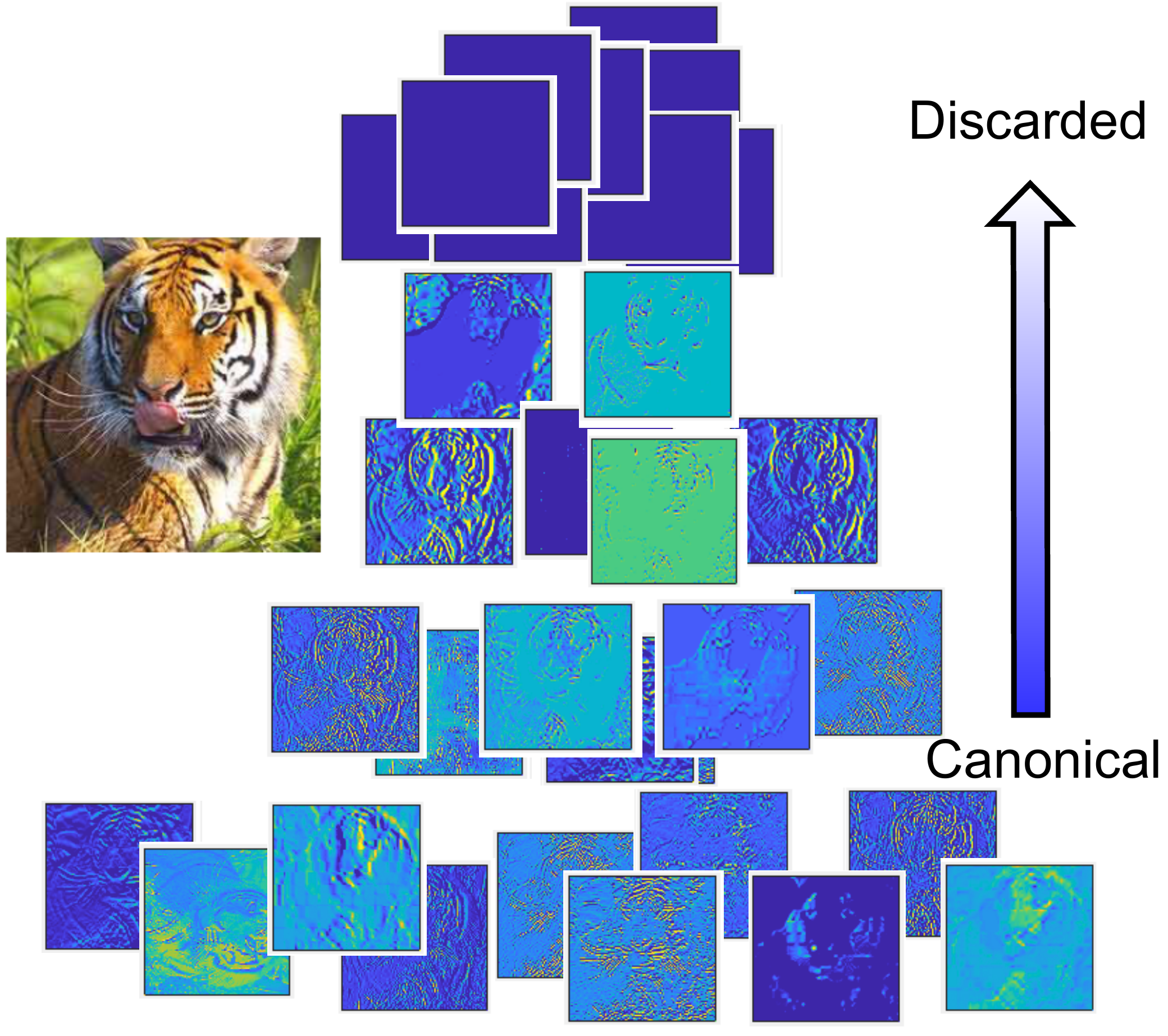}
    \caption{MSGC is different form channel pruning, where the rank of channels is simply described by ``pruned'' and ``unpruned''. The group mechanism allows to create a richer pyramid structure: on the bottom are the canonical or fundamental channels that are selected by all groups, then the less general but potentially more specific channels for particular groups when we looking upward, finally the discarded channels on the top that are completely pruned for the current image sample.}
    \label{fig:pyramid}
\end{figure}

It is worth mentioning that the above convolution process is different to the normal convolution, the standard grouped convolution, and network pruning. First, compared with the normal convolution where the 5-8th output channels are generated by all the input channels, convolution in MSGC only chooses a part of input channels which saves computation. Second, instead of choosing the input channels in a predefined way in the standard grouped convolution (\eg, the last two input channels will be chosen as the input for group 2 in standard grouped convolution), MSGC applies a mask generator to dynamically choose the channels, which brings more learnability. Finally, compared with network pruning, MSGC preservers more input channels. In other words, the input channels that are not selected (or pruned) by one group may be selected by other groups. We illustrate those specific chosen input channels for a random image sample in \cref{fig:pyramid}, where most of the channels are chosen by at least one group. Therefore, the channels are not easily pruned (please also see the following discussion).


\vspace{0.3em}
\noindent \textbf{Step 3: Decoupling gating and attention (Optional).} We can attach a parallel MLP to generate another saliency tensor $\mathbf{A}_i^{G_i\times C_i}\in\mathbb{R}^{G_i\times C_i}$ as the attention tensor for layer $L_i$. The input feature maps in the selected channels will then be rescaled with the associated attention values.



\vspace{0.3em}
\noindent \textbf{End-to-end learning.} We may notice the only non-differentiable part is the Sign function, which converts the continuous saliency tensors to binary gating masks. However, we could use gradient approximation techniques like the straight-through estimator (STE)~\cite{bengio2013ste} or the Gumbel-Softmax reparameterization technique~\cite{jang2016gumbelsoftmax} such that the whole pipeline can be optimized by gradient descent in an end-to-end manner. 
Once the gating masks are learnable with no restrictions \emph{w.r.t.} groups, layers, and samples, the ``middle-spectrum'' space is freely unfolded.

\subsection{Discussions}

\subsubsection{Pyramid structure of feature maps}

It is interesting to find that the group mechanism helps to rank the input feature maps in a softer way. Specifically, channel pruning only learns to select important channels once. \emph{I.e.}, once channels have been pruned, they are totally discarded. Grouped convolution preserves all channels but reduces the computation by conducting convolution in groups. When channel pruning and grouped convolution come together in MSGC, channels are not easily discarded because every group has an independent selection process. As shown in \cref{fig:pyramid}, we can identify those channels that are selected by all groups (canonical channels), channels selected by part of the groups (less important but still valuable channels), and finally the discarded channels (meaningless channels). The importance scores of channels become more reliable, leading to better performance compared with channel pruning-based methods (\cref{tab:table-channel-pruning}).

\subsubsection{Discussion about plug-in}

MSGC serves as a plug-in module for a backbone for reducing its computational cost and retaining its predictive performance to a large extent. In the implementation, the filters of a normal convolutional layer $W\in \mathbb{R}^{k\times k\times C_i\times C_{i+1}}$ can be divided to $\{W_g\in \mathbb{R}^{k\times k\times C_i\times \frac{C_{i+1}}{G_i}}|g=1,2,...,G_i\}$ as the initial state of the filters in each group when the module is plugged in.

\subsubsection{Discussion about computational cost}

The reduction of computational cost of a network block, where MSGC is applied, mainly comes from the sparse connections in the groups. It is also attributed to the fact that the gating masks for all layers are generated using the input of $L_1$ of the block. In other words, the group topology (formulated as $\mathbf{B}_i^{G_i\times C_i}$) for the whole block can be already determined given the input of the first layer. If a certain channel is not selected by any group in a later layer in the block, we can directly skip the calculation of this channel in the previous layer. For example, if the 2nd input channel of $L_3$ is not selected by any group in $L_3$, we can directly skip the convolutions in the previous layer $L_2$ that generate this channel. Besides, the MACs of the light MLPs is negligible compared with the original model. For example, in ResNet-18, MLPs take 1.8 million MACs, which is less than $0.1\%$ of that of the original model that has 1.8 billion MACs. 

\subsection{Loss function}
MACs of the network can be calculated based on the binary gating masks $\mathbf{B}_i^{G_i\times C_i}$. A zero value in the gating masks indicates a certain part of convolutions are skipped. Take the same example in the illustration of Step 1 of MSGC, $\mathbf{B}_3^{2,5}=0$ indicates the 5th input channel of $L_3$ is not considered as an input for calculating the 2nd group output channels, its associated MACs (=$k\times k\times \frac{C_4}{G_3}$, where $k$ is the kernel size) are therefore not included in the final MACs. 

To control both the final MACs and accuracy of the model during training, we adopts a budget loss $\mathcal{L}_{bgt}$ and the standard task-specific loss $\mathcal{L}_{task}$ (\eg, cross entropy loss for image classification). Specifically, the budget loss is formulated as:
\begin{equation}
    \mathcal{L}_{bgt} = \text{max}(\lambda (\frac{\mathcal{M}_b}{\mathcal{M}_{ori}} - \tau), 0),
\label{eq:loss}
\end{equation}
where $\mathcal{M}_b$ is the mean running MACs over the current mini-batch, $\mathcal{M}_{ori}$ is the MACs of the original model, $\lambda$ is a controlling parameter, and $\tau$ is a target remaining rate. A more detailed description about $\lambda$ and $\tau$ can be found in the next section. The final loss is $\mathcal{L} = \mathcal{L}_{bgt} + \mathcal{L}_{task}$.


\section{Experiments}
\label{sec:experiments}
To demonstrate the effectiveness of the proposed method, we compare it with prior state-of-the-art (SOTA) methods including both channel pruning- and grouped convolution-based ones. The generalizability of our method is also validated on the task of object detection under both one- and two-stage frameworks. Model analysis and ablation study is given in the last part of this section.

\subsection{Image classification}
\label{sec:image_classificatoin}

\noindent \textbf{Dataset}: We use the large scale ImageNet dataset~\cite{imagenet} to validate MSGC.
There are 1.2 million images for training and 50K images for validation with 1K classes. During training, the data is augmented with random cropping (to size $224 \times 224$) and random horizontal flipping. The Top-1 and Top-5 accuracy on validation set are used for comparison.

\vspace{0.3em}
\noindent \textbf{Training setup}: As a plug-in module, MSGC has the advantage to directly prune and boost the pretrained models. We start by setting $\tau$ (in \cref{eq:loss}) to $1.0$, then progressively decrease it to a target value $\tau_{end}$ during the first half of the training process. Precisely, if we want to reduce MACs of the network by half, we set $\tau_{end} = 0.5$. By using a big $\lambda$ (\emph{i.e.}, $\lambda = 30$ in the experiments), we gradually reduce the computational cost to a target budget (\ie, $\tau_{end}\cdot \mathcal{M}_{ori}$). The second half of training can be seen as a fine-tuning process. Following~\cite{su2020dgc}, we train ResNet-18/50 and CondenseNet for 120 epochs and MobileNetV2 for 150 epochs, with a batch size of 256. The initial learning rate is set to 0.075 for the MLPs, and 0.015 for the pretrained weights in the backbone, which are both decayed with cosine-shape annealing~\cite{huang2018condensenet} to 0. Stochastic gradient descent (SGD) with momentum of 0.9 is used as the optimizer, weight decay is set to $10^{-4}$ and only applied on the backbone weights. We use Gumbel-Softmax~\cite{jang2016gumbelsoftmax} when binarizing the saliency vectors (please refer to appendix A for more details). The whole process is implemented with the Pytorch library~\cite{paszke2019pytorch} on two NVIDIA A100 GPUs. 

\vspace{0.3em}
\noindent \textbf{Backbones}: To make a comprehensive comparison with prior approaches, we choose the widely adopted backbones, namely, ResNet-18/50, DenseNet, and MobileNetV2. Our MSGC module is plugged in those backbones with the configurations illustrated in \cref{tab:setting}.

\begin{table}[t!]
\caption{MSGC configurations on different backbones. Attention layers means to which layers in the block the attention (step 3) MLPs are attached.}
\centering
\setlength{\tabcolsep}{0.008\linewidth}
\resizebox*{\linewidth}{!}{
\begin{tabular}{llcc}
\toprule
Backbone & Block type & $\{G_i|i=1, 2, ..., M\}$ & Attention layers \\
\midrule
ResNet-18~\cite{he2016resnet} & Basic block & $\{1, 4\}$ & 1, 2 \\
ResNet-50~\cite{he2016resnet} & Bottleneck block & $\{1, 4, 1\}$ & 2 \\
DenseNet~\cite{huang2017densenet} & Condensenet block~\cite{huang2018condensenet} & $\{1, 4\}$ & 2 \\
MobileNetV2~\cite{sandler2018mobilenetv2} & Reverted Bottleneck block & $\{1, 4\}$ & 2 \\
\bottomrule
\end{tabular}
}
\label{tab:setting}
\end{table}

\begin{table*}[t!]
\caption{Comparison with prior SOTA \textbf{channel pruning}-based methods on ResNet-18, ResNet-50, and MobileNetV2 backbones. For MAC, the underlined numbers are from the original papers and we calculated the rest. The accuracy is record in a style of \emph{baseline} $\to$ \emph{pruned}. $\Delta$MAC$\downarrow$ means MAC reduction rate, the larger, the more the model is pruned.}
\centering
\setlength{\tabcolsep}{0.012\linewidth}
\resizebox*{\linewidth}{!}{
\begin{tabular}{l|c|cc|cc|c|c}
\toprule
Backbone & Method & Top-1 (\%) & $\Delta$Top-1 (\%) & Top-5 (\%) & $\Delta$Top-5 (\%) & MAC & $\Delta$MAC$\downarrow$ (\%)\\
\midrule
\multirow{12}{*}{ResNet-18} & SFP~\cite{DBLP:sfp} & 70.28 $\to$ 67.10 & -3.18 & 89.63 $\to$ 87.78 & -1.85 & 1.06B & \underline{\emph{41.8}}\\
& DCP~\cite{DBLP:dcp} & 69.64 $\to$ 67.35 & -2.29 & 88.98 $\to$ 87.60 & -1.38 & 0.91B & \underline{\emph{50.0}}\\
& FPGM~\cite{he2019fpgm} & 70.28 $\to$ 68.41 & -1.87 & 89.63 $\to$ 88.48 & -1.15 & 1.06B & \underline{\emph{41.8}}\\
& DSA~\cite{ning2020dsa} & 69.72 $\to$ 68.61 & -1.11 & 89.07 $\to$ 88.35 & -0.72 & 1.09B & \underline{\emph{40.0}}\\
& PFP~\cite{DBLP:pfp} & 69.74 $\to$ 65.65 & -4.09 & 89.07 $\to$ 86.75 & -2.32 & 1.03B & \underline{\emph{43.1}}\\
& FBS~\cite{DBLP:fbs} & 70.71 $\to$ 68.17 & -2.54 & 89.68 $\to$ 88.22 & -1.46 & 0.92B & \underline{\emph{49.5}}\\
& DRL-based~\cite{DBLP:drlbased} & 69.76 $\to$ 68.73 & -1.03 & 89.08 $\to$ 88.65 & -0.43 & 0.94B & \underline{\emph{48.5}}\\
& ManiDP~\cite{tang2021manidp} & 69.76 $\to$ 68.88 & -0.88 & 89.08 $\to$ 88.76 & -0.32 & 0.89B & \underline{\emph{51.0}}\\
& FFP~\cite{lee2023fastfilterpruning} & 69.45 $\to$ 68.85 & -0.60 & 88.86 $\to$ 88.47 & -0.39 & 1.09B & \underline{40.0} \\
& DNCP~\cite{zheng2022dncp} & 70.10 $\to$ 69.20 & -0.90 & - & - & \underline{1.04B} & 42.8 \\
\cmidrule(r){2-8}
& \textbf{MSGC w/o Attention} & \textbf{69.76 $\to$ 70.30} & \textbf{+0.54} & \textbf{89.08 $\to$ 89.27} & \textbf{+0.19} & \textbf{0.88B} & \textbf{51.4}\\
& \textbf{MSGC} & \textbf{69.76 $\to$ 71.51} & \textbf{+1.75} & \textbf{89.08 $\to$ 90.21} & \textbf{+1.13} & \textbf{0.88B} & \textbf{51.3}\\
\midrule
\multirow{17}{*}{ResNet-50} & SFP~\cite{DBLP:sfp} & 76.15 $\to$ 74.61 & -1.54 & 92.87 $\to$ 92.06 & -0.81 & 2.38B & \underline{\emph{41.8}}\\
& DCP~\cite{DBLP:dcp} & 76.01 $\to$ 74.95 & -1.06 & 92.93 $\to$ 92.32 & -0.61 & 2.05B & \underline{\emph{50.0}}\\
& FPGM~\cite{he2019fpgm} & 76.15 $\to$ 75.59 & -0.56 & 92.87 $\to$ 92.63 & -0.24 & 2.37B & \underline{\emph{42.2}}\\
& MetaPruning~\cite{liu2019metapruning} & 76.60 $\to$ 75.40 & -1.20 & - & - & \underline{\emph{2.00B}} & 51.1\\
& DMC~\cite{gao2020dmc} & 76.15 $\to$ 75.35 & -0.80 & 92.87 $\to$ 92.49 & -0.38 & 1.84B & \underline{\emph{55.0}}\\
& LeGR~\cite{chin2020legr} & 76.10 $\to$ 75.70 & -0.40 & 92.90 $\to$ 92.70 & -0.20 & 2.37B & \underline{\emph{42.0}}\\
& DSA~\cite{ning2020dsa} & 76.02 $\to$ 74.69 & -1.33 & 92.86 $\to$ 92.06 & -0.80 & 2.05B & \underline{\emph{50.0}}\\
& PFP~\cite{DBLP:pfp} & 76.13 $\to$ 75.21 & -0.92 & 92.87 $\to$ 92.43 & -0.45 & 2.86B & \underline{\emph{30.1}}\\
& DPFPS~\cite{ruan2021dpfps} & 76.15 $\to$ 75.55 & -0.60 & 92.87 $\to$ 92.54 & -0.33 & 2.20B & \underline{\emph{46.2}}\\
& NPPM~\cite{gao2021nppm} & 76.15 $\to$ 75.96 & -0.19 & 92.87 $\to$ 92.75 & -0.12 & \underline{\emph{1.81B}} & \underline{\emph{56.0}}\\
& DS-Net~\cite{li2021dsnet} & 76.10 $\to$ 76.10 & 0 & - & - & \underline{\emph{2.20B}} & 46.2\\
& FFP~\cite{lee2023fastfilterpruning} & 76.18 $\to$ 76.23 & +0.05 & 92.79 $\to$ 92.87 & +0.08 & \underline{2.48B} & 39.5 \\
 & ASTER~\cite{zhang2023adaptivefilterpruning} & 76.15 $\to$ 75.52 & -0.63 & 92.87 $\to$ 92.65 & -0.22 & \underline{1.75B} & \underline{57.3} \\
& DNCP~\cite{zheng2022dncp} & 76.60 $\to$ 76.30 & -0.30 & - & - & \underline{2.20B} & 46.3 \\
& White-Box~\cite{zhang2022whitebox} & 76.15 $\to$ 75.32 & -0.83 & 92.96 $\to$ 92.43 & -0.53 & \underline{2.22B} & \underline{45.6} \\
\cmidrule(r){2-8}
& \textbf{MSGC w/o Attention} & \textbf{76.13 $\to$ 77.20} & \textbf{+1.07} & \textbf{92.86 $\to$ 93.37} & \textbf{+0.51} & \textbf{1.89B} & \textbf{54.0}\\
& \textbf{MSGC} & \textbf{76.13 $\to$ 76.76} & \textbf{+0.63} & \textbf{92.86 $\to$ 92.99} & \textbf{+0.13} & \textbf{1.89B} & \textbf{53.8}\\
\midrule
\multirow{11}{*}{MobileNetV2} & DCP~\cite{DBLP:dcp} & 70.11 $\to$ 64.13 & -5.89 & - & -3.77 & 169M & \underline{\emph{44.8}}\\
& MetaPruning~\cite{liu2019metapruning} & 72.00 $\to$ 71.20 & -0.80 & - & - & \underline{\emph{217M}} & 29.3\\
& DMC~\cite{gao2020dmc} & 71.88 $\to$ 68.37 & -3.51 & 90.29 $\to$ 88.46 & -1.83 & 166M & \underline{\emph{46.0}}\\
& LeGR~\cite{chin2020legr} & 71.80 $\to$ 71.40 & -0.40 & - & - & 215M & \underline{\emph{30.0}}\\
& DPFPS~\cite{ruan2021dpfps} & 72.00 $\to$ 71.10 & -0.90 & - & - & 231M & \underline{\emph{24.9}}\\
& ManiDP~\cite{tang2021manidp} & 71.80 $\to$ 71.42 & -0.38 & 90.43 $\to$ 90.28 & -0.15 & 193M & \underline{\emph{37.2}}\\
& NPPM~\cite{gao2021nppm} & 72.00 $\to$ 72.02 & +0.02 & 90.38 $\to$ 90.26 & -0.12 & \underline{\emph{221M}} & \underline{\emph{29.7}}\\
& ASTER~\cite{zhang2023adaptivefilterpruning} & 72.00 $\to$ 71.89 & -0.11 & - & - & \underline{221M} & \underline{29.7} \\
& DNCP~\cite{zheng2022dncp} & 72.30 $\to$ 72.70 & +0.40 & - & - & \underline{211M} & 31.3 \\
\cmidrule(r){2-8}
& \textbf{MSGC w/o Attention} & \textbf{71.88 $\to$ 72.10} & \textbf{+0.22} &\textbf{ 90.27 $\to$ 90.41} & \textbf{+0.14} & \textbf{198M} & \textbf{35.4}\\
& \textbf{MSGC} & \textbf{71.88 $\to$ 72.59} & \textbf{+0.71} & \textbf{90.27 $\to$ 90.82} & \textbf{+0.55} & \textbf{197M} & \textbf{35.8}\\
\bottomrule
\end{tabular}
}
\label{tab:table-channel-pruning}
\end{table*}

\begin{table*}[t!]
\caption{Comparison with prior SOTA \textbf{grouped convolution}-based methods on ResNet and MobileNetV2 backbones. The results of MobileNetV2 (0.7) are cited from IGCV3~\cite{DBLP:igcv3}. Top-1 (\%) is recorded as \emph{baseline} $\to$ \emph{compressed} and $\Delta$MAC$\downarrow$ (\%) means MAC reduction.}
\centering
\setlength{\tabcolsep}{0.008\linewidth}
\resizebox*{\linewidth}{!}{
\begin{tabular}{lccc|ccc|ccc}
\toprule
Method & \multicolumn{3}{c}{ResNet-18} & \multicolumn{3}{c}{ResNet-50} & \multicolumn{3}{c}{MobileNetV2}\\
\midrule
Top-1 (\%) & $\Delta$Top-1 (\%) & $\Delta$MAC$\downarrow$ (\%) & Top-1 (\%) & $\Delta$Top-1 (\%) & $\Delta$MAC$\downarrow$ (\%) & Top-1 (\%) & $\Delta$Top-1 (\%) & $\Delta$MAC$\downarrow$ (\%)\\
\midrule
CGNet~\cite{DBLP:CGNet} & 69.20 $\to$ 68.80 & -0.40 & 48.2 & - & - & - & - & - & - \\
Deep Roots~\cite{ioannou2017deeproots} & - & - & - & 73.00 $\to$ 73.20 & +0.20 & 48.4 & - & - & - \\
SG-CNN~\cite{guo2020sgc} & - & - & - & 76.13 $\to$ 75.20 & -0.93 & 53.3 & - & - & - \\
MobileNetV2 (0.7) & - & - & - & - & - & - & 71.30 $\to$ 66.51 & -4.79 & 20.2 \\
IGCV3~\cite{DBLP:igcv3} & - & - & - & - & - & - & 71.30 $\to$ 68.45 & -2.85 & 31.6 \\
DGC~\cite{su2020dgc} & 69.76 $\to$ 68.78 & -0.98 & 51.0 & - & - & - & 72.00 $\to$ 70.70 & -1.30 & 20.2 \\
\midrule
\textbf{MSGC w/o Attention} & \textbf{69.76 $\to$ 70.30} & \textbf{+0.54} & \textbf{51.4} & \textbf{76.13 $\to$ 77.20} & \textbf{+1.07} & \textbf{54.0} & \textbf{71.88 $\to$ 72.10} & \textbf{+0.22} & \textbf{35.4} \\
\textbf{MSGC} & \textbf{69.76 $\to$ 71.51} & \textbf{+1.75} & \textbf{51.3} & \textbf{76.13 $\to$ 76.76} & \textbf{+0.63} & \textbf{53.8} & \textbf{71.88 $\to$ 72.59} & \textbf{+0.71} & \textbf{35.8} \\
\bottomrule
\end{tabular}
}
\label{tab:table-grouped-convolution}
\end{table*}

In \cref{tab:table-channel-pruning} and \cref{tab:table-grouped-convolution}, we compare MSGC with the prior SOTA works based on channel pruning and grouped convolution respectively on both ResNet and MobileNetV2 backbones. Due to the fact that there is a relatively lack of works on DenseNet, we compare MSGC with the previous methods that aimed to build light DenseNet-like structures, namely, the CondenseNet~\cite{huang2018condensenet} variants, by replacing the LGC~\cite{huang2018condensenet} module in CondenseNet with MSGC (the same for other modules in other methods). The results on different methods based on CondenseNet are given in \cref{tab:table-densenet}.

Unlike prior methods, MSGC maintains accuracy with a more aggressive MAC reduction when other prior methods often cause accuracy drop. Notably, we even obtain accuracy gains for ResNet and MobileNetV2 when the MAC is considerably reduced by 50\% and 35\% respectively. Comparing with the latest method FFP~\cite{lee2023fastfilterpruning}, our method improves ResNet-18 by 1.75\% with 51.3\% MAC reduction, while FFP decreases the accuracy by 0.60\% with a smaller MAC reduction rate (40.0\%). On ResNet-50, our method increases the accuracy by 0.63\% with a considerable 53.8\% MAC reduction, while FFP only has a 0.05\% accuracy increase, even with a smaller MAC reduction rate (39.5\%). We conjecture that the phenomenon may result from the reorganization of channel usage from two perspectives: first, the effective removal of redundant parameters; second, the more reliable ranking of input channels as shown in \cref{fig:pyramid} (also see our analysis in \cref{sec:ablation}).

We also completely remove the attention MLPs in our models, denoted as ``MSGC w/o Attention'' in the tables, to see the influence of our attention-wise design.
Generally, the attention-wise design works well on relatively light backbones like ResNet-18 and MobileNetV2, but may give diminishing or negative returns for large backbones like DenseNet and ResNet-50, meaning that a saturated status is reached under current configurations before applying attentions. 

\begin{figure}[t!]
\centering
    \centering
    \includegraphics[width=\linewidth]{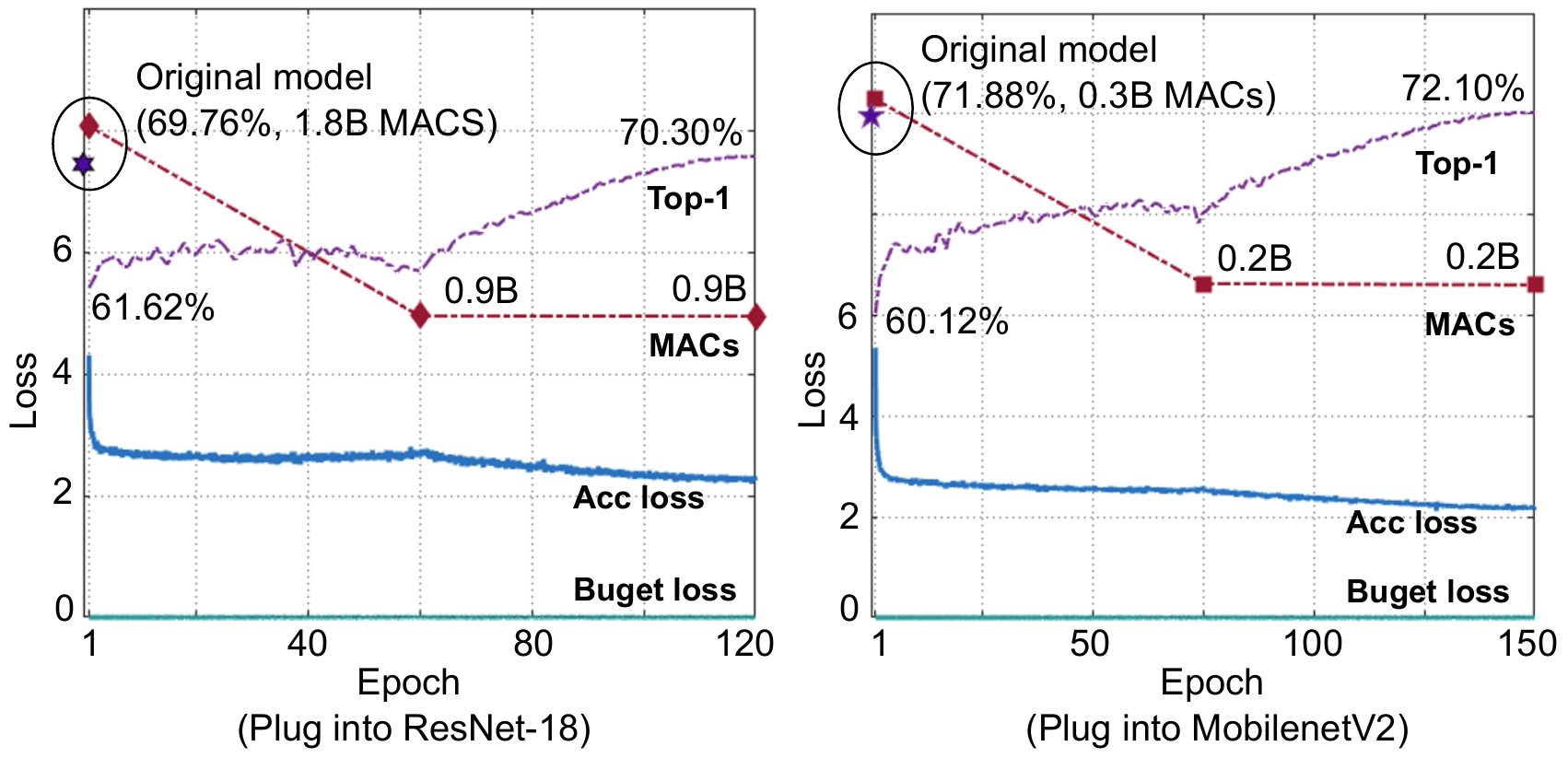}
    \caption{Curves of validation accuracy, computational cost, and training losses during the training process.}
    \label{fig:logs}
\end{figure}

An illustration of training dynamics is shown in \cref{fig:logs}, where we could see the computational cost stably drops to the budget with a slowly growing accuracy in the first half of the training, with the budget loss keeping around 0. In the second half of training, the accuracy is significantly compensated.

\begin{table*}[t!]
\caption{Experimental results on MS COCO dataset. We calculate the average running MACs of the backbone over the validation set.}
\centering
\setlength{\tabcolsep}{0.008\linewidth}
\resizebox*{\linewidth}{!}{
\begin{tabular}{lcccc|lcccc}
\toprule
\multicolumn{5}{c}{Faster-RCNN} & \multicolumn{5}{c}{RetinaNet}\\
\midrule
Backbone & mAP (\%) & AP$_{50}$ (\%) & AP$_{70}$ (\%) & MAC & Backbone & mAP (\%) & AP$_{50}$ (\%) & AP$_{70}$ (\%) & MAC\\
\midrule
ResNet-50 & 38.1 & 59.3 & 41.0 & 74.1B & ResNet-50 & 36.1 & 55.5 & 38.4 & 74.1B\\
ResNet-50-MSGC & 37.4 & 59.1 & 40.2 & 33.8B & ResNet-50-MSGC & 36.0 & 56.1 & 38.0 & 34.2B\\
\midrule
MobileNetV2 & 32.7 & 53.4 & 34.7 & 1.4B & MobileNetV2 & 30.7 & 48.7 & 32.6 & 1.4B\\
MobileNetV2-0.75 & 30.4 & 50.8 & 31.8 & 1.0B & MobileNetV2-0.75 & 28.5 & 46.1 & 29.6 & 1.0B\\
MobileNetV2-MSGC & 32.1 & 53.2 & 33.6 & 0.9B & MobileNetV2-MSGC & 31.3 & 50.1 & 32.9 & 0.9B\\
\bottomrule
\end{tabular}
}
\label{tab:table-coco}
\end{table*}

\begin{table}[t!]
\caption{Comparison with prior SOTA methods on building the CondenseNet architecture. The proposed method achieves similar or superior performance with less MACs.}
\centering
\setlength{\tabcolsep}{0.02\linewidth}
\resizebox*{\linewidth}{!}{
\begin{tabular}{lccc}
\toprule
Method & Top-1 (\%) & Top-5 (\%) & MAC\\
\midrule
CondenseNet (standard GC) & 71.0 & 90.1 & 529M\\
CondenseNet-LGC~\cite{huang2018condensenet} & 73.8 & 91.7 & 529M\\
CondenseNet-FLGC~\cite{wang2019flgc} & 74.7 & 92.1 & 529M\\
CondenseNet-DGC~\cite{su2020dgc} & 74.6 & 92.2 & 549M\\
CondenseNet-OGC~\cite{li2022ogcnet} & 74.6 & - & 529M \\
\midrule
CondenseNet-MSGC & 74.8 & 92.2 & 523M\\
\bottomrule
\end{tabular}
}
\label{tab:table-densenet}
\end{table}

\subsection{Object detection}

\noindent \textbf{Dataset}: We conduct our experiments on the COCO 2017 detection dataset~\cite{lin2014mscoco}, which has 118K and 5K images for training and validation. During training, the images are resized with the longer edge $=$ 1333, combined with random horizontal flipping. During validation, the images are resized such that the shorter edge $\leq 800$, and the longer edge $\leq 1333$.

\vspace{0.3em}
\noindent \textbf{Backbones}: We evaluate MSGC on ResNet-50 and MobileNetV2 under Faster R-CNN~\cite{ren2015fasterrcnn} (2-stage) and RetinaNet~\cite{lin2017retinanet} (1-stage) frameworks. The configurations of the backbones are the same as in \cref{sec:image_classificatoin}. Observed from the previous section on image classification, we attach the attention MLPs to MobileNetV2 and not to ResNet-50.

\vspace{0.3em}
\noindent \textbf{Training setup}:
The training scheme mostly follows the experiments on ImageNet. The differences are illustrated as follows. All the models are trained for 90,000 iterations with 16 images in a mini-batch. The basic initial learning rate is set as 0.02 for Faster-RCNN and 0.01 for RetinaNet. For the weights in the backbone (including the MLPs in the MSGC module), the initial learning rate is further multiplied with 0.5. Learning rate is decayed with a factor of 0.1 at iteration 60,000 and 80,000. For others like batch size per image, we adopted the default settings in the Detectron2 library~\cite{wu2019detectron2}. The whole process is implemented with the Pytorch library~\cite{paszke2019pytorch} on two NVIDIA A100 GPUs.

The results are shown in \cref{tab:table-coco}. For ResNet-50, MSGC can effectively cut the MACs of the backbone by $54\%$ with only a slight or negligible accuracy drop. For MobileNetV2, MSGC outperforms MobileNetV2-0.75 by a large margin with a similar computational overhead on both frameworks. Notably, on the RetinaNet framework, the performance of MobileNetV2 can also be improved with the plugged MSGC, from 30.7\% to 31.3\% in mAP, with a $28.5\%$ reduction on MACs. The comparison shows MSGC has a good generalizability on the object detection task.

\subsection{Model analysis and ablation study}
\label{sec:ablation}

In this section, we give our analysis from the dimensions of group-wise, layer-wise, sample-wise, and attention-wise, respectively. Experiments are based on \cref{sec:image_classificatoin}. More exploratory studies are given at the end. The attention MLPs are not applied in MSGC for simplicity if not specified.

\begin{table}[t!]
\caption{Influence of group numbers.}
\centering
\setlength{\tabcolsep}{0.02\linewidth}
\resizebox*{\linewidth}{!}{
\begin{tabular}{l|c|cc|c}
\toprule
Backbone & Group & Top-1 ($\%$) & Top-5 ($\%$) & MAC\\
\midrule
\multirow{5}{*}{ResNet-18} & baseline & 69.76 & 89.08 & 1.82B\\
\cmidrule(r){2-5}
& 1 & 69.72 & 89.12 & 0.89B\\
& 2 & 70.02 & 89.21 & 0.88B\\
& 4 & 70.30 & 89.27 & 0.88B\\
& 8 & 70.10 & 89.25 & 0.88B\\
\midrule
\multirow{5}{*}{MobileNetV2} & baseline & 71.88 & 90.27 & 307M\\
\cmidrule(r){2-5}
& 1 & 71.58 & 90.28 & 199M\\
& 2 & 71.98 & 90.38 & 199M\\
& 4 & 72.10 & 90.41 & 198M\\
& 8 & 71.84 & 90.19 & 198M\\
\bottomrule
\end{tabular}
}
\label{tab:ablation-group}
\end{table}

\vspace{0.3em}
\noindent \textbf{Group-wise}: 
Configuration of a MSGC module can be varied. Though in \cref{sec:image_classificatoin}, we empirically set the group numbers $\{G_i\}$, it is definitely possible to seek more advanced techniques like neural architecture search (NAS)~\cite{elsken2019nas} to give a better structural setting. Here, we changed the group numbers on ResNet-18 (more specifically, we change the group number for the 2nd layer of the Basic block) and MobileNetV2 under similar MACs. The results are shown in \cref{tab:ablation-group}.

\begin{figure}[t!]
\centering
    \centering
    \includegraphics[width=\linewidth]{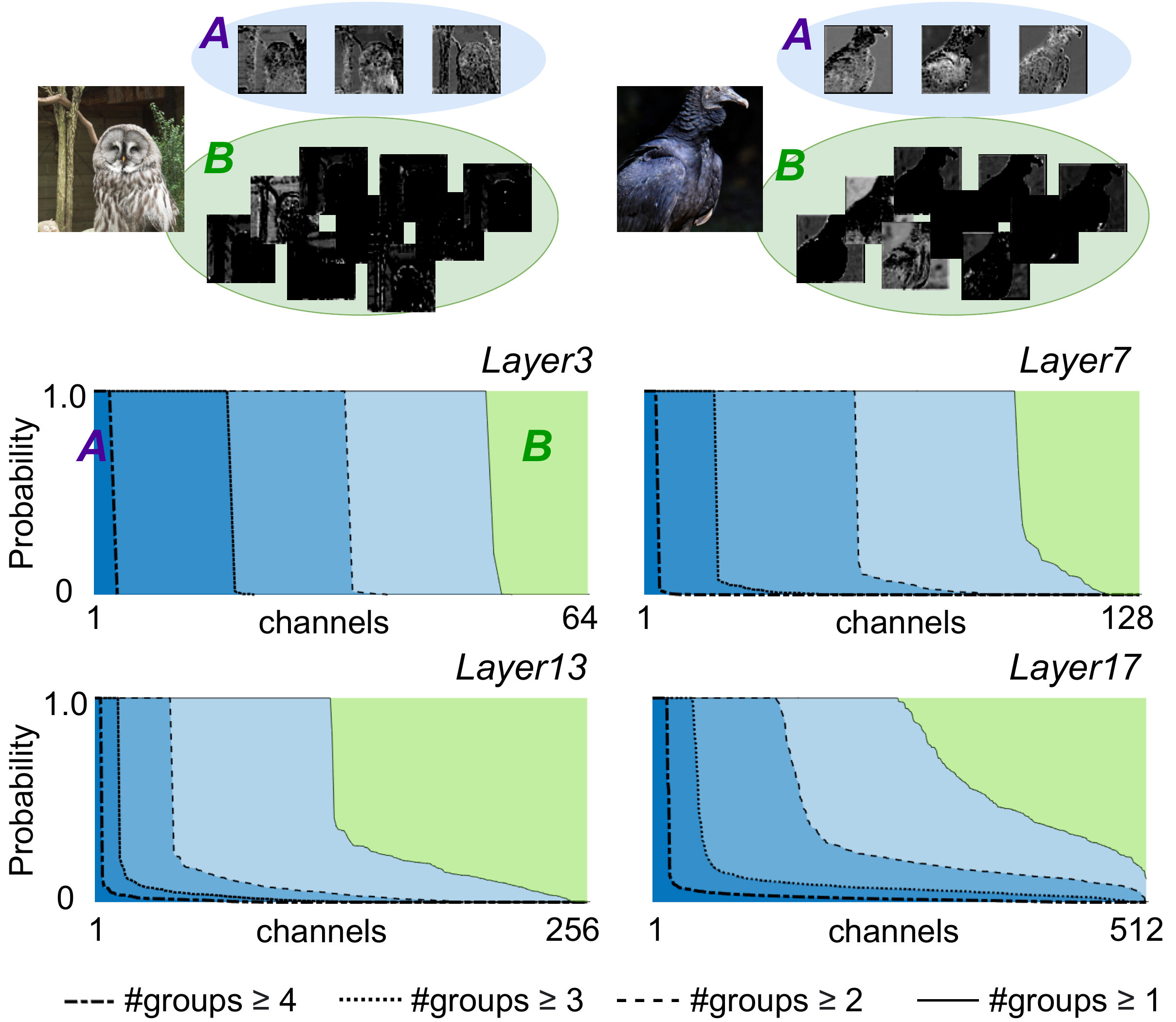}
    \caption{In each layer, we calculated the probability for each input channel of being selected by at least $g$ groups over the ImageNet validation set, based on ResNet-18. Each curve was drawn by ranking the probabilities in descending order. In each sub-figure in the bottom part, curves from left to right represents $g=\{4,3,2,1\}$ respectively. The ``A'' area definitely represents the canonical channels that are constantly used by all the groups. In contrast, ``B'' represents the discarded channels that might be meaningless or redundant.}
    \label{fig:groupwise}
\end{figure}

\begin{figure}[t!]
\centering
    \centering
    \includegraphics[width=0.9\linewidth]{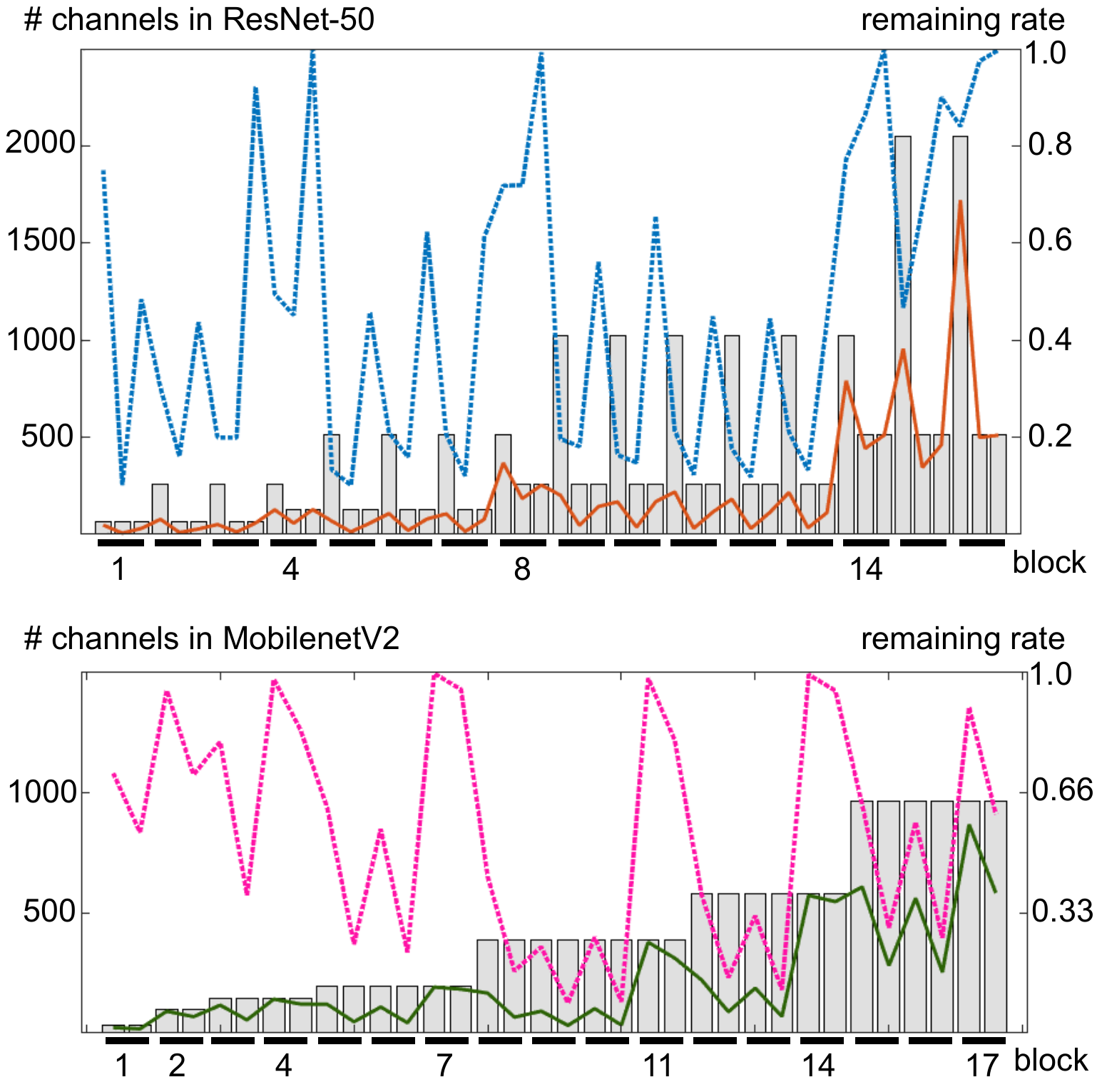}
    \caption{Illustrations of the pruned structures. The gray bars show the number of input channels in each layer of the original models. The middle orange or green line indicates the average number of channels used in each group. The top blue or red line represents the remaining rate (the result of the value in the middle line divided by the value of the gray bar, a higher value means more channels are remained).}
    \label{fig:layerwise}
\end{figure}

Group $=1$ reduces MSGC to a dynamic channel pruning module. Like~\cite{DBLP:fbs}, the dynamic execution helps the network to prune redundant information specific to the input samples, potentially leading to better discrimination ability. As expected, using groups achieves an increasing improvement. It peaks at group $=4$ since a bigger group number may bring extra learning complexity~\cite{su2020dgc}.

We dig deeper into the pyramid-style reuse of the feature channels, as illustrated in \cref{fig:groupwise}. MSGC can detect those canonical, general, and meaningless channels automatically, leading to a more reliable channel ranking. It is worth mentioning that the curves in shallower layers look steeper (means a major proportion of samples selected the same channels), which is in line with the fact that feature maps in shallow layers usually contain low-level or fundamental image information shared by most samples. While in deeper layers, samples become more distinguished due to their varying semantic characteristics.

\begin{figure}[t!]
\centering
    \centering
    \includegraphics[width=0.95\linewidth]{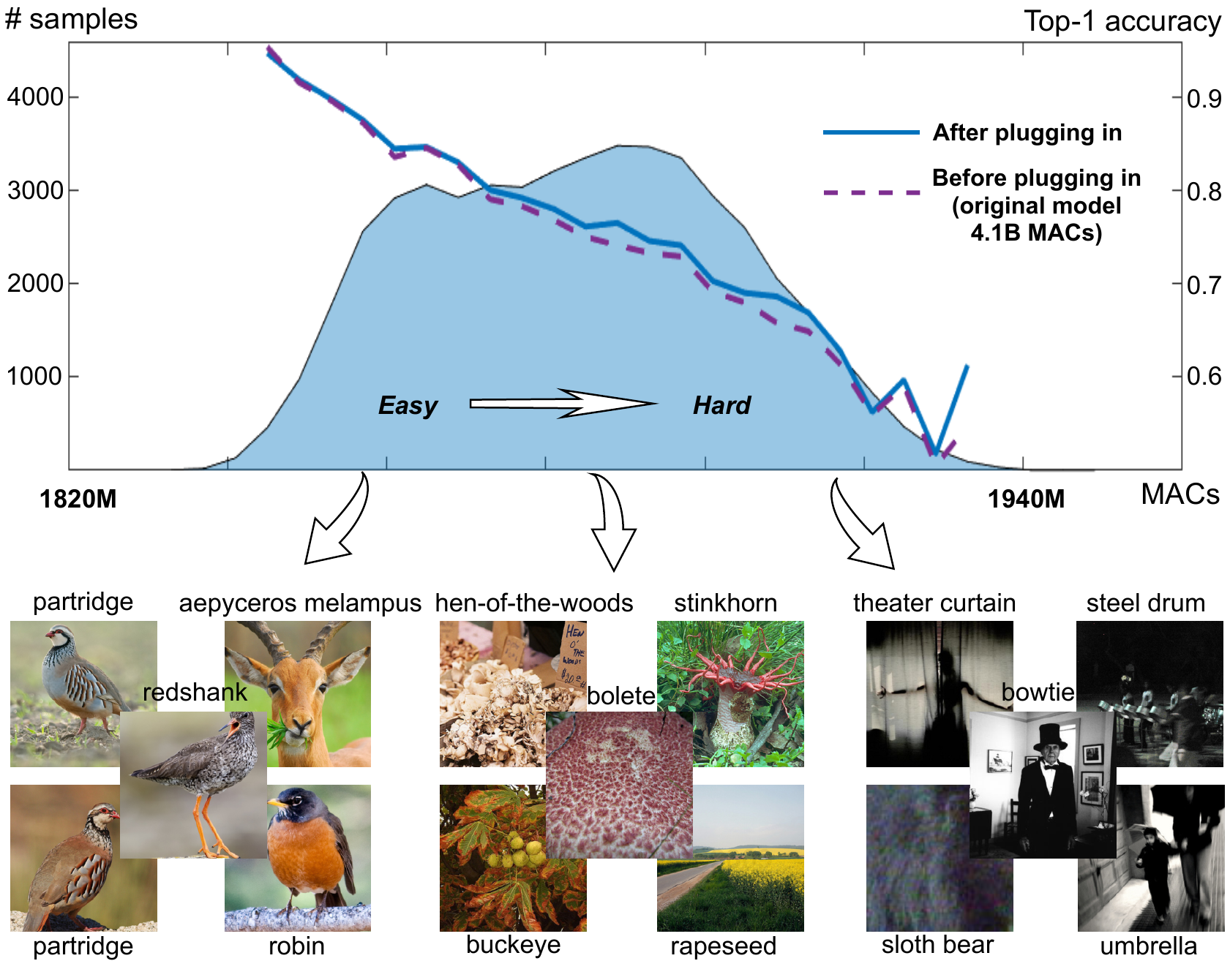}
    \caption{The histogram of MACs over the ImageNet validation set. The prediction accuracy curves are also plotted for the original (dashed purple line) and the MSGC equipped (solid blue line) model respectively, based on the ResNet-50 architecture. Each value in the accuracy curves represents the corresponding Top-1 accuracy of the sample set falling in the current histogram bin.}
    \label{fig:samplewise}
\end{figure}

\vspace{0.3em}
\noindent \textbf{Layer-wise}: 
We illustrate the dynamically pruned structures in \cref{fig:layerwise} for ResNet-50 and MobileNetV2, where we get some consistent observations on both backbones similar to~\cite{liu2019metapruning}. Specifically, the significant peaks of the remaining rate curve appear on the layers where there is a downsampling operation of the image. Since the reduction of resolution shrink the information in a feature map, more channels are needed in order to preserve the capacity for the next layers. In ResNet-50, there are also some secondary peaks in the last layer of each block, which expand the channels for the next Bottleneck block. Therefore, it is expected to carry more input channels to hold the information.

\begin{figure}[t!]
\centering
    \centering
    \includegraphics[width=\linewidth]{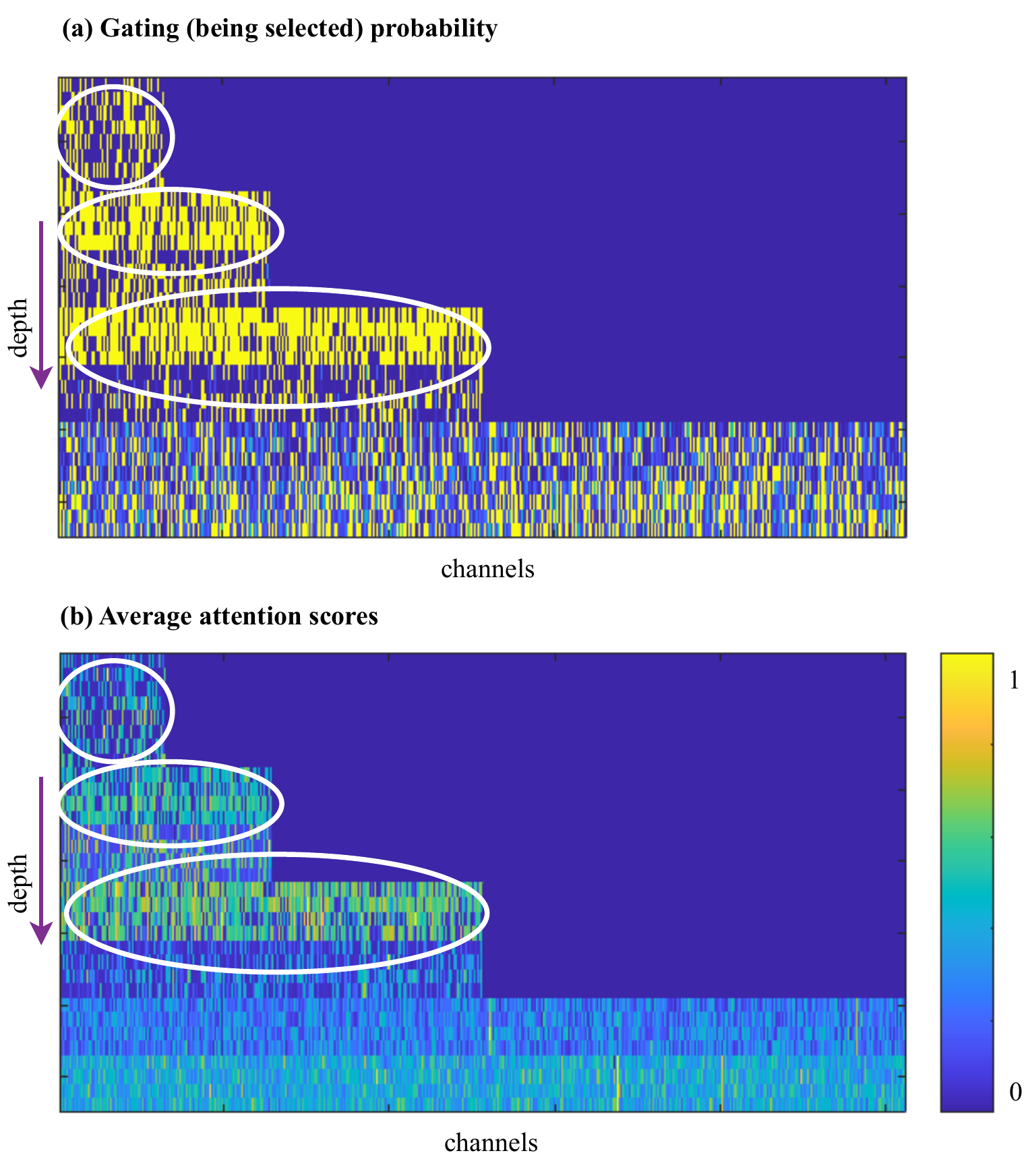}
    \caption{The visualization is based on ResNet-18, where each row represents a group in a certain layer, every four rows represent a convolutional layer. Therefore, each element in the visualized matrices indicates an input channel for a certain group in a certain layer. Left is the the gating probability of the channels, calculated based on the associated saliency values over the ImageNet validation set. Right is the attached average attention values.}
    \label{fig:attentionwise}
\end{figure}

\vspace{0.3em}
\noindent \textbf{Sample-wise}:
As mentioned in \cref{sec:method}, MSGC gives an adaptive resource allocation due to its dynamic property. This is shown in \cref{fig:samplewise}. It is found that both the original model and the MSGC equipped model can obtain good prediction accuracies on those ``easy'' samples. However, MSGC can automatically spend more computation on those ``harder'' samples, leading to a better prediction (remembering that the original model uses much more MACs on each sample equally). Therefore, the overall performance can be enhanced with MSGC. Interestingly, the ``easy'' and ``hard'' samples determined by MSGC are quite intuitive for humans, which are partially illustrated at the bottom of \cref{fig:samplewise}. For example, ``easy'' samples have less cluttered backgrounds and easy to be recognized by human eyes, while the ``hard'' samples show the opposite (\emph{e.g.}, the rightmost samples even tend to be in a black and white style and more confusing).

\vspace{0.3em}
\noindent \textbf{Attention-wise}: Consistent with \cref{fig:groupwise}, in \cref{fig:attentionwise} (a), shallower layers show a relatively fixed gating pattern, \ie, the gating probabilities (the probabilities of being selected) in shallow layers are either close to zero or close to one. The close-to-one values indicate the associated channels (the yellow cells) were selected by a major part of samples. While deeper layers have larger variations. It is interesting to find that the visualized attention matrix in \cref{fig:attentionwise} (b) shows a similar sparse pattern with the gating matrix. In other words, the channels with close-to-zero gating probabilities also have close-to-zero attentions. However, different to the gating probabilities in shallow layers, the attention scores in shallow layers have more middle-range values, which inject extra variations that can enrich the representation of those layers. 

\begin{table}[t!]
\caption{Influence of $\lambda$ on the performance on ImageNet.}
\centering
\setlength{\tabcolsep}{0.008\linewidth}
\resizebox*{\linewidth}{!}{
\begin{tabular}{l|cc|cc}
\toprule
$\lambda$ & Target MAC ($\tau_{end} = 0.5$) & Actual MAC & Top-1 ($\%$) & Top-5 ($\%$)\\
\hline
baseline & - & 1.82B & 69.76 & 89.08\\
\hline
1 & 0.89B & 1.21B & 71.66 & 90.18\\
10 & 0.89B & 0.88B & 70.13 & 89.25\\
30 & 0.89B & 0.88B & 70.30 & 89.27\\
60 & 0.89B & 0.88B & 70.18 & 89.35\\
\bottomrule
\end{tabular}
}
\label{tab:lambda}
\end{table}


\vspace{0.3em}
\noindent \textbf{Influence of the controlling parameter $\lambda$}: We conduct the experiments with ResNet-18 on ImageNet dataset to show how the controlling parameter $\lambda$, which controls the weight of the budget loss (\cref{eq:loss}) during training, influences the performance. The results are shown in \cref{tab:lambda}. We can see that a too small $\lambda$ may cause a larger model than expected. When $\lambda$ is big enough (\emph{i.e.}, $\geq 30$), we can precisely control the target MAC and the method is quite stable.


\begin{table}[t!]
\caption{Exploration of model accuracy with different budgets (by changing the value of $\tau_{end}$ to control the final MAC).}
\centering
\setlength{\tabcolsep}{0.025\linewidth}
\resizebox*{\linewidth}{!}{
\begin{tabular}{l|cccc}
\toprule
$\tau_{end}$ & MAC & $\Delta$MAC$\downarrow$ (\%) & Top-1 (\%) & $\Delta$Top-1 (\%) \\
\midrule
- & 1.82B & - & 69.76 & - \\
0.3 & 0.54B & 70.1 & 65.88 & -3.88\\
0.5 & 0.88B & 51.4 & 70.30 & +0.54 \\
0.7 & 1.26B & 30.6 & 71.50 & +1.74 \\
0.9 & 1.63B & 10.3 & 72.33 & +2.57 \\
0.99 & 1.76B & 0.3 & 72.20 & +2.44 \\
\bottomrule
\end{tabular}
}
\label{tab:balance}
\end{table}

\vspace{0.3em}
\noindent \textbf{From MAC reducer to accuracy enhancer}:
Setting different values of $\tau_{end}$ controls the computational overhead of the final model. For example, it is demonstrated in the previous sections that $\tau_{end}=0.5$ leads to a computational reduction of 50\% on the ResNet structures without sacrificing the accuracy. In this study, we give a more detailed exploration of how MSGC balances accuracy and MAC by changing $\tau_{end}$. The results are shown in \cref{tab:balance} where ResNet-18 is used. Since $\tau_{end}<1$, MSGC always acts as a ``MAC reducer'' for the original baseline architecture. A small $\tau_{end}$ results in a good computation reduction, but at the same time a degraded performance on prediction accuracy. However, when the computation reduction is not our focus, we can simply apply MSGC as an effective accuracy enhancer. Surprisingly, when $\tau_{end}=0.9$, the model achieves a absolute 2.57\% Top-1 accuracy gain, from 69.76\% to 72.33\%. Meanwhile, further increasing $\tau_{end}$ to 0.99 does not incur additional accuracy gain, implying necessary removal of redundant or meaningless features is important for achieving the optimal performance.

\begin{table}[t!]
\caption{The inference time is calculated only for the convolutional operations, averaged over 1000 samples in the ImageNet validation set, with an Intel i7-8700 device.}
\centering
\setlength{\tabcolsep}{0.03\linewidth}
\resizebox*{\linewidth}{!}{
\begin{tabular}{l|cccc}
\toprule
Backbone & MSGC & MAC & Top-1 ($\%$) & Latency\\
\hline
\multirow{2}{*}{ResNet-18} & \xmark & 1.82B & 69.76 & 12.10ms\\
& \checkmark & 0.88B & 70.30 & 8.94ms\\
\hline
\multirow{2}{*}{ResNet-50} & \xmark & 4.10B & 76.13 & 26.46ms\\
& \checkmark & 1.89B & 77.20 & 19.02ms\\
\bottomrule
\end{tabular}
}
\label{tab:latency}
\end{table}

\vspace{0.3em}
\noindent \textbf{Discussion on hardware implementation}:
When our focus is to reduce computation, one factor we may consider is the actual hardware implementation. It is indeed a common challenge for dynamic execution networks to achieve actual acceleration in real-world implementation, since the current deep learning hardware and libraries were mostly optimized for static models~\cite{han2021dynamicsurvey}. Specifically, extra burden of indexing and weight copying usually leads to significantly inefficient computation waste as investigated in~\cite{li2021dsnet}. In addition, dynamic networks usually couple an extra controller to tell which part of the main backbone should be executed~\cite{su2020dgc,DBLP:fbs,lin2017rnp} (like the MLPs in ours), which also hinders parallel implementation. Therefore, many of those works have been tailored together with hardware to obtain the practical efficiency, including the adaptive channel pruning ones~\cite{DBLP:CGNet,akhlaghi2018snapea}. 
The most related work to ours regarding hardware implementation is CGNet~\cite{DBLP:CGNet}. Both adopt grouped convolution and involve dynamic execution. CGNet even introduced a finer sparsity granularity (at pixel level), which  relies on a highly customized implementation on the ASIC accelerator, and achieved a speed-up close to the theoretical one, while worked worse on the common GPU or CPU. In contrast, MSGC conducts dynamic execution on the channels rather than pixels, hence should be easy to get the actual acceleration as well with a tailored design in the target hardware. In \cref{tab:latency}, we also give a reference of efficiency by comparing the latency only on the convolutional operations in the backbones with and without MSGC.

\section{Conclusion}
\label{sec:conclusion}
In this paper, we proposed MSGC which explores the broad ``middle spectrum'' area between channel pruning and grouped convolution. MSGC benefits from both worlds and acts as a plug-in module for various backbones like ResNet, DenseNet, and MobileNetV2. Particularly, MSGC can reduce the computational costs of those backbones by from $35\%$ to $50\%$, without sacrificing accuracy on the large scale ImageNet dataset. Besides, MSGC shows good generalizability when applied to the object detection task. Finally, MSGC can also be used as an accuracy booster without increasing the computational cost.


\appendix

\subsection{Binarizing saliency vectors to binary gating masks}

Supposing the binarization process aims to convert the saliency vector $\mathbf{x}\in \mathbb{R}^{1\times C}$ to a binary vector $\mathbf{x}^b\in \{0, 1\}^{1\times C}$, we do the binarization element-wise, \emph{i.e.}, for each element $x$ in $\mathbf{x}$, we convert it to a binary value $x^b$. As mentioned in the main paper, we can use the following Sign function to efficiently binarize $x$ during testing:
\begin{equation}
    \text{Sign}(x) = \begin{cases}
    1\;\;\text{if}\;\; x \geq 0\\
    0\;\;\text{otherwise},
    \end{cases}\\
\end{equation}

While during training, the Sign function is non-differentiable. In this way, $x^b$ can be seen as a binary random variable with the probabilities $P(x^b=1) = \pi_1 = \sigma(x)$ and $P(x^b=0) = \pi_{0} = 1 - \pi_1$, where $\sigma$ is the Sigmoid function
\begin{equation}
    \sigma(x) = \frac{1}{1+e^{-x}}.
\end{equation}

We use the Gumbel-Softmax~\cite{jang2016gumbelsoftmax} reparameterization trick to recalculate the probability distribution as:
\begin{equation}
    P(x^b = i) = \frac{\exp((\log(\pi_i) + g_i)/\tau)}{\sum_{j=\{0,1\}}\exp((\log(\pi_j) + g_j)/\tau)},
\end{equation}
where $g_i$ is a random variable sampled from a Gumbel distribution and $\tau$ is the temperature (set as $2/3$ in the implementation suggested in~\cite{dgnet}).
Therefore, 
\begin{align}
    &P(x^b = 1)\nonumber\\ &=\frac{\exp((\log(\pi_1) + g_1)/\tau)}{\exp((\log(\pi_0) + g_0)/\tau) + \exp((\log(\pi_1) + g_1)/\tau)}\nonumber\\
    &=\frac{1}{1+\frac{\exp((\log(\pi_0) + g_0)/\tau)}{\exp((\log(\pi_1) + g_1)/\tau)}}\nonumber\\
    &=\frac{1}{(1-\exp(\log(\pi_1) - \log(\pi_0) + g_1 - g_0)/\tau)}\nonumber\\
    &=\sigma((\log(\frac{\pi_1}{1-\pi_1}) + g_1 - g_0)/\tau).
\end{align}
Because $\pi_1 = \sigma(x)$, it turns out 
\begin{align}
    &P(x^b = 1)\nonumber\\
    &=\sigma((x+g_1-g_0)/\tau).
\end{align}
Since the difference of $g_0$ and $g_1$ follows a Logistic distribution, sampling from $g_1-g_0$ can be implemented by sampling from $\log(v/(1-v))$ where $v$ follows the normal distribution.
The training strategy is:
\begin{equation}
x^b = \begin{cases}
    \text{Sign}(P(x^b=1)-0.5)\;\;&\text{(forward)},\\
    P(x^b=1)\;\;&\text{(backward)}.
    \end{cases}\\
\end{equation}

\bibliographystyle{IEEEtran}
\bibliography{IEEEabrv,msgc}

\begin{thebibliography}{10}
\providecommand{\url}[1]{#1}
\csname url@samestyle\endcsname
\providecommand{\newblock}{\relax}
\providecommand{\bibinfo}[2]{#2}
\providecommand{\BIBentrySTDinterwordspacing}{\spaceskip=0pt\relax}
\providecommand{\BIBentryALTinterwordstretchfactor}{4}
\providecommand{\BIBentryALTinterwordspacing}{\spaceskip=\fontdimen2\font plus
\BIBentryALTinterwordstretchfactor\fontdimen3\font minus
  \fontdimen4\font\relax}
\providecommand{\BIBforeignlanguage}[2]{{%
\expandafter\ifx\csname l@#1\endcsname\relax
\typeout{** WARNING: IEEEtran.bst: No hyphenation pattern has been}%
\typeout{** loaded for the language `#1'. Using the pattern for}%
\typeout{** the default language instead.}%
\else
\language=\csname l@#1\endcsname
\fi
#2}}
\providecommand{\BIBdecl}{\relax}
\BIBdecl

\bibitem{wang2021sodsurvey}
W.~Wang, Q.~Lai, H.~Fu, J.~Shen, H.~Ling, and R.~Yang, ``Salient object
  detection in the deep learning era: An in-depth survey,'' \emph{IEEE
  Transactions on Pattern Analysis and Machine Intelligence}, vol.~44, no.~6,
  pp. 3239--3259, 2021.

\bibitem{su2021pdc}
Z.~Su, W.~Liu, Z.~Yu, D.~Hu, Q.~Liao, Q.~Tian, M.~Pietikainen, and L.~Liu,
  ``Pixel difference networks for efficient edge detection,'' in \emph{ICCV},
  2021, pp. 5117--5127.

\bibitem{li2017improving}
Y.~Li, Y.~Song, and J.~Luo, ``Improving pairwise ranking for multi-label image
  classification,'' in \emph{CVPR}, 2017, pp. 3617--3625.

\bibitem{liu2020objectdetection}
L.~Liu, W.~Ouyang, X.~Wang, P.~Fieguth, J.~Chen, X.~Liu, and
  M.~Pietik{\"a}inen, ``Deep learning for generic object detection: A survey,''
  \emph{International Journal of Computer Vision}, vol. 128, no.~2, pp.
  261--318, 2020.

\bibitem{chen2021anatomy}
T.~Chen, C.~Fang, X.~Shen, Y.~Zhu, Z.~Chen, and J.~Luo, ``Anatomy-aware 3d
  human pose estimation with bone-based pose decomposition,'' \emph{IEEE
  Transactions on Circuits and Systems for Video Technology}, vol.~32, no.~1,
  pp. 198--209, 2021.

\bibitem{tan2019efficientnet}
M.~Tan and Q.~Le, ``Efficientnet: Rethinking model scaling for convolutional
  neural networks,'' in \emph{ICML}.\hskip 1em plus 0.5em minus 0.4em\relax
  PMLR, 2019, pp. 6105--6114.

\bibitem{lee2023fastfilterpruning}
S.~Lee and B.~C. Song, ``Fast filter pruning via coarse-to-fine neural
  architecture search and contrastive knowledge transfer,'' \emph{IEEE
  Transactions on Neural Networks and Learning Systems}, 2023.

\bibitem{zhang2023adaptivefilterpruning}
Y.~Zhang and N.~M. Freris, ``Adaptive filter pruning via sensitivity
  feedback,'' \emph{IEEE Transactions on Neural Networks and Learning Systems},
  2023.

\bibitem{howard2017mobilenets}
A.~G. Howard, M.~Zhu, B.~Chen, D.~Kalenichenko, W.~Wang, T.~Weyand,
  M.~Andreetto, and H.~Adam, ``Mobilenets: Efficient convolutional neural
  networks for mobile vision applications,'' \emph{arXiv preprint
  arXiv:1704.04861}, 2017.

\bibitem{sandler2018mobilenetv2}
M.~Sandler, A.~Howard, M.~Zhu, A.~Zhmoginov, and L.-C. Chen, ``Mobilenetv2:
  Inverted residuals and linear bottlenecks,'' in \emph{CVPR}, 2018, pp.
  4510--4520.

\bibitem{zhang2022dynamicthreshold}
J.~Zhang, Z.~Su, Y.~Feng, X.~Lu, M.~Pietik{\"a}inen, and L.~Liu, ``Dynamic
  binary neural network by learning channel-wise thresholds,'' in \emph{ICASSP
  2022-2022 IEEE International Conference on Acoustics, Speech and Signal
  Processing (ICASSP)}.\hskip 1em plus 0.5em minus 0.4em\relax IEEE, 2022, pp.
  1885--1889.

\bibitem{su2022svnet}
Z.~Su, M.~Welling, M.~Pietik{\"a}inen, and L.~Liu, ``Svnet: Where so (3)
  equivariance meets binarization on point cloud representation,'' in
  \emph{2022 International Conference on 3D Vision (3DV)}.\hskip 1em plus 0.5em
  minus 0.4em\relax IEEE, 2022, pp. 547--556.

\bibitem{zhang2022whitebox}
Y.~Zhang, M.~Lin, C.-W. Lin, J.~Chen, Y.~Wu, Y.~Tian, and R.~Ji, ``Carrying out
  cnn channel pruning in a white box,'' \emph{IEEE Transactions on Neural
  Networks and Learning Systems}, 2022.

\bibitem{luo2017thinet}
J.-H. Luo, J.~Wu, and W.~Lin, ``Thinet: A filter level pruning method for deep
  neural network compression,'' in \emph{ICCV}, 2017, pp. 5058--5066.

\bibitem{he2017channelpruing}
Y.~He, X.~Zhang, and J.~Sun, ``Channel pruning for accelerating very deep
  neural networks,'' in \emph{ICCV}, 2017, pp. 1389--1397.

\bibitem{han2016deepcompression}
S.~Han, H.~Mao, and W.~J. Dally, ``Deep compression: Compressing deep neural
  network with pruning, trained quantization and huffman coding,'' in
  \emph{ICLR}, 2016.

\bibitem{chen2021sharinggroup}
T.~Chen, B.~Duan, Q.~Sun, M.~Zhang, G.~Li, H.~Geng, Q.~Zhang, and B.~Yu, ``An
  efficient sharing grouped convolution via bayesian learning,'' \emph{IEEE
  Transactions on Neural Networks and Learning Systems}, vol.~33, no.~12, pp.
  7367--7379, 2021.

\bibitem{su2020dgc}
Z.~Su, L.~Fang, W.~Kang, D.~Hu, M.~Pietik{\"a}inen, and L.~Liu, ``Dynamic group
  convolution for accelerating convolutional neural networks,'' in
  \emph{ECCV}.\hskip 1em plus 0.5em minus 0.4em\relax Springer, 2020, pp.
  138--155.

\bibitem{xie2017aggregated}
S.~Xie, R.~Girshick, P.~Doll{\'a}r, Z.~Tu, and K.~He, ``Aggregated residual
  transformations for deep neural networks,'' in \emph{CVPR}, 2017, pp.
  1492--1500.

\bibitem{ioannou2017deeproots}
Y.~Ioannou, D.~Robertson, R.~Cipolla, and A.~Criminisi, ``Deep roots: Improving
  cnn efficiency with hierarchical filter groups,'' in \emph{CVPR}, 2017, pp.
  1231--1240.

\bibitem{DBLP:igcv3}
K.~Sun, M.~Li, D.~Liu, and J.~Wang, ``{IGCV3:} interleaved low-rank group
  convolutions for efficient deep neural networks,'' in \emph{BMVC}, 2018, p.
  101.

\bibitem{zhang2018shufflenet}
X.~Zhang, X.~Zhou, M.~Lin, and J.~Sun, ``Shufflenet: An extremely efficient
  convolutional neural network for mobile devices,'' in \emph{CVPR}, 2018, pp.
  6848--6856.

\bibitem{zhang2017igcv1}
T.~Zhang, G.-J. Qi, B.~Xiao, and J.~Wang, ``Interleaved group convolutions,''
  in \emph{ICCV}, 2017, pp. 4373--4382.

\bibitem{xie2018igcv2}
G.~Xie, J.~Wang, T.~Zhang, J.~Lai, R.~Hong, and G.-J. Qi, ``Interleaved
  structured sparse convolutional neural networks,'' in \emph{CVPR}, 2018, pp.
  8847--8856.

\bibitem{zhao2019unitary}
R.~Zhao, Y.~Hu, J.~Dotzel, C.~D. Sa, and Z.~Zhang, ``Building efficient deep
  neural networks with unitary group convolutions,'' in \emph{CVPR}, 2019, pp.
  11\,303--11\,312.

\bibitem{gao2020channelnets}
H.~Gao, Z.~Wang, L.~Cai, and S.~Ji, ``Channelnets: Compact and efficient
  convolutional neural networks via channel-wise convolutions,'' \emph{IEEE
  Transactions on Pattern Analysis and Machine Intelligence}, 2020.

\bibitem{zhang2023scgnet}
H.~Zhang, S.~Lai, Y.~Wang, Z.~Da, Y.~Dun, and X.~Qian, ``Scgnet: Shifting and
  cascaded group network,'' \emph{IEEE Transactions on Circuits and Systems for
  Video Technology}, 2023.

\bibitem{szegedy2015inceptionv1}
C.~Szegedy, W.~Liu, Y.~Jia, P.~Sermanet, S.~Reed, D.~Anguelov, D.~Erhan,
  V.~Vanhoucke, and A.~Rabinovich, ``Going deeper with convolutions,'' in
  \emph{CVPR}, 2015, pp. 1--9.

\bibitem{szegedy2016inceptionv2}
C.~Szegedy, V.~Vanhoucke, S.~Ioffe, J.~Shlens, and Z.~Wojna, ``Rethinking the
  inception architecture for computer vision,'' in \emph{CVPR}, 2016, pp.
  2818--2826.

\bibitem{chollet2017xception}
F.~Chollet, ``Xception: Deep learning with depthwise separable convolutions,''
  in \emph{CVPR}, 2017, pp. 1251--1258.

\bibitem{huang2018condensenet}
G.~Huang, S.~Liu, L.~Van~der Maaten, and K.~Q. Weinberger, ``Condensenet: An
  efficient densenet using learned group convolutions,'' in \emph{CVPR}, 2018,
  pp. 2752--2761.

\bibitem{zheng2022dncp}
Y.-J. Zheng, S.-B. Chen, C.~H. Ding, and B.~Luo, ``Model compression based on
  differentiable network channel pruning,'' \emph{IEEE Transactions on Neural
  Networks and Learning Systems}, 2022.

\bibitem{he2019fpgm}
Y.~He, P.~Liu, Z.~Wang, Z.~Hu, and Y.~Yang, ``Filter pruning via geometric
  median for deep convolutional neural networks acceleration,'' in \emph{CVPR},
  2019, pp. 4340--4349.

\bibitem{he2016resnet}
K.~He, X.~Zhang, S.~Ren, and J.~Sun, ``Deep residual learning for image
  recognition,'' in \emph{CVPR}, 2016, pp. 770--778.

\bibitem{huang2017densenet}
G.~Huang, Z.~Liu, L.~Van Der~Maaten, and K.~Q. Weinberger, ``Densely connected
  convolutional networks,'' in \emph{CVPR}, 2017, pp. 4700--4708.

\bibitem{imagenet}
J.~Deng, W.~Dong, R.~Socher, L.-J. Li, K.~Li, and L.~Fei-Fei, ``Imagenet: A
  large-scale hierarchical image database,'' in \emph{CVPR}, 2009, pp.
  248--255.

\bibitem{lin2014mscoco}
T.-Y. Lin, M.~Maire, S.~Belongie, J.~Hays, P.~Perona, D.~Ramanan,
  P.~Doll{\'a}r, and C.~L. Zitnick, ``Microsoft coco: Common objects in
  context,'' in \emph{ECCV}.\hskip 1em plus 0.5em minus 0.4em\relax Springer,
  2014, pp. 740--755.

\bibitem{DBLP:conf/iclr/l1norm}
H.~Li, A.~Kadav, I.~Durdanovic, H.~Samet, and H.~P. Graf, ``Pruning filters for
  efficient convnets,'' in \emph{ICLR}, 2017.

\bibitem{DBLP:sfp}
Y.~He, G.~Kang, X.~Dong, Y.~Fu, and Y.~Yang, ``Soft filter pruning for
  accelerating deep convolutional neural networks,'' in \emph{IJCAI}, 2018, pp.
  2234--2240.

\bibitem{DBLP:dcp}
Z.~Zhuang, M.~Tan, B.~Zhuang, J.~Liu, Y.~Guo, Q.~Wu, J.~Huang, and J.~Zhu,
  ``Discrimination-aware channel pruning for deep neural networks,'' in
  \emph{NeurIPS}, 2018, pp. 883--894.

\bibitem{wang2019tmm-pruning}
Z.~Wang, W.~Hong, Y.-P. Tan, and J.~Yuan, ``Pruning 3d filters for accelerating
  3d convnets,'' \emph{IEEE Transactions on Multimedia}, vol.~22, no.~8, pp.
  2126--2137, 2019.

\bibitem{ruan2021dpfps}
X.~Ruan, Y.~Liu, B.~Li, C.~Yuan, and W.~Hu, ``Dpfps: Dynamic and progressive
  filter pruning for compressing convolutional neural networks from scratch,''
  in \emph{AAAI}, vol.~35, no.~3, 2021, pp. 2495--2503.

\bibitem{liu2019metapruning}
Z.~Liu, H.~Mu, X.~Zhang, Z.~Guo, X.~Yang, K.-T. Cheng, and J.~Sun,
  ``Metapruning: Meta learning for automatic neural network channel pruning,''
  in \emph{ICCV}, 2019, pp. 3296--3305.

\bibitem{ople2021tmm-pruning2}
J.~J.~M. Ople, T.-M. Huang, M.-C. Chiu, Y.-L. Chen, and K.-L. Hua, ``Adjustable
  model compression using multiple genetic algorithms,'' \emph{IEEE
  Transactions on Multimedia}, 2021.

\bibitem{DBLP:conf/iclr/slimmable}
J.~Yu, L.~Yang, N.~Xu, J.~Yang, and T.~S. Huang, ``Slimmable neural networks,''
  in \emph{ICLR}, 2019.

\bibitem{li2021dsnet}
C.~Li, G.~Wang, B.~Wang, X.~Liang, Z.~Li, and X.~Chang, ``Dynamic slimmable
  network,'' in \emph{CVPR}, 2021, pp. 8607--8617.

\bibitem{DBLP:fbs}
X.~Gao, Y.~Zhao, L.~Dudziak, R.~D. Mullins, and C.~Xu, ``Dynamic channel
  pruning: Feature boosting and suppression,'' in \emph{ICLR}, 2019.

\bibitem{tang2021manidp}
Y.~Tang, Y.~Wang, Y.~Xu, Y.~Deng, C.~Xu, D.~Tao, and C.~Xu, ``Manifold
  regularized dynamic network pruning,'' in \emph{CVPR}, 2021, pp. 5018--5028.

\bibitem{NIPS2012_alexnet}
A.~Krizhevsky, I.~Sutskever, and G.~E. Hinton, ``Imagenet classification with
  deep convolutional neural networks,'' in \emph{NeurIPS}, F.~Pereira, C.~J.~C.
  Burges, L.~Bottou, and K.~Q. Weinberger, Eds., vol.~25.\hskip 1em plus 0.5em
  minus 0.4em\relax Curran Associates, Inc., 2012.

\bibitem{wang2019flgc}
X.~Wang, M.~Kan, S.~Shan, and X.~Chen, ``Fully learnable group convolution for
  acceleration of deep neural networks,'' in \emph{CVPR}, 2019, pp. 9049--9058.

\bibitem{zhang2019dgconv}
Z.~Zhang, J.~Li, W.~Shao, Z.~Peng, R.~Zhang, X.~Wang, and P.~Luo,
  ``Differentiable learning-to-group channels via groupable convolutional
  neural networks,'' in \emph{ICCV}, 2019, pp. 3542--3551.

\bibitem{guo2020sgc}
Q.~Guo, X.-J. Wu, J.~Kittler, and Z.~Feng, ``Self-grouping convolutional neural
  networks,'' \emph{Neural Networks}, vol. 132, pp. 491--505, 2020.

\bibitem{wu2018blockdrop}
Z.~Wu, T.~Nagarajan, A.~Kumar, S.~Rennie, L.~S. Davis, K.~Grauman, and
  R.~Feris, ``Blockdrop: Dynamic inference paths in residual networks,'' in
  \emph{CVPR}, 2018, pp. 8817--8826.

\bibitem{veit2018convnetaig}
A.~Veit and S.~Belongie, ``Convolutional networks with adaptive inference
  graphs,'' in \emph{CVPR}, 2018, pp. 3--18.

\bibitem{hu2018seblock}
J.~Hu, L.~Shen, and G.~Sun, ``Squeeze-and-excitation networks,'' in
  \emph{CVPR}, 2018, pp. 7132--7141.

\bibitem{yang2019condconv}
B.~Yang, G.~Bender, Q.~V. Le, and J.~Ngiam, ``Condconv: Conditionally
  parameterized convolutions for efficient inference,'' in \emph{NeurIPS},
  2019, pp. 1305--1316.

\bibitem{dgnet}
F.~Li, G.~Li, X.~He, and J.~Cheng, ``Dynamic dual gating neural networks,'' in
  \emph{ICCV}, 2021.

\bibitem{verelst2020dynamicconvlutionpixel}
T.~Verelst and T.~Tuytelaars, ``Dynamic convolutions: Exploiting spatial
  sparsity for faster inference,'' in \emph{CVPR}, 2020, pp. 2320--2329.

\bibitem{DBLP:conf/iclr/decomposing}
Y.~Li, Y.~Chen, X.~Dai, M.~Liu, D.~Chen, Y.~Yu, L.~Yuan, Z.~Liu, M.~Chen, and
  N.~Vasconcelos, ``Revisiting dynamic convolution via matrix decomposition,''
  in \emph{ICLR}, 2021.

\bibitem{chen2020dynamicrelu}
Y.~Chen, X.~Dai, M.~Liu, D.~Chen, L.~Yuan, and Z.~Liu, ``Dynamic relu,'' in
  \emph{ECCV}.\hskip 1em plus 0.5em minus 0.4em\relax Springer, 2020, pp.
  351--367.

\bibitem{huang2018msdn}
G.~Huang, D.~Chen, T.~Li, F.~Wu, L.~van~der Maaten, and K.~Weinberger,
  ``Multi-scale dense networks for resource efficient image classification,''
  in \emph{ICLR}, 2018.

\bibitem{wang2020glance}
Y.~Wang, K.~Lv, R.~Huang, S.~Song, L.~Yang, and G.~Huang, ``Glance and focus: a
  dynamic approach to reducing spatial redundancy in image classification,'' in
  \emph{NeurIPS}, 2020.

\bibitem{han2021dynamicsurvey}
Y.~Han, G.~Huang, S.~Song, L.~Yang, H.~Wang, and Y.~Wang, ``Dynamic neural
  networks: A survey,'' \emph{IEEE Transactions on Pattern Analysis and Machine
  Intelligence}, 2021.

\bibitem{DBLP:dst}
J.~Liu, Z.~Xu, R.~Shi, R.~C.~C. Cheung, and H.~K. So, ``Dynamic sparse
  training: Find efficient sparse network from scratch with trainable masked
  layers,'' in \emph{ICLR}, 2020.

\bibitem{ioffe2015bn}
S.~Ioffe and C.~Szegedy, ``Batch normalization: Accelerating deep network
  training by reducing internal covariate shift,'' in \emph{ICML}.\hskip 1em
  plus 0.5em minus 0.4em\relax PMLR, 2015, pp. 448--456.

\bibitem{bengio2013ste}
Y.~Bengio, N.~L{\'e}onard, and A.~Courville, ``Estimating or propagating
  gradients through stochastic neurons for conditional computation,''
  \emph{arXiv preprint arXiv:1308.3432}, 2013.

\bibitem{jang2016gumbelsoftmax}
E.~Jang, S.~Gu, and B.~Poole, ``Categorical reparameterization with
  gumbel-softmax,'' in \emph{ICLR}, 2017.

\bibitem{paszke2019pytorch}
A.~Paszke, S.~Gross, F.~Massa, A.~Lerer, J.~Bradbury, G.~Chanan, T.~Killeen,
  Z.~Lin, N.~Gimelshein, L.~Antiga \emph{et~al.}, ``Pytorch: An imperative
  style, high-performance deep learning library,'' in \emph{NeurIPS}, 2019, pp.
  8026--8037.

\bibitem{ning2020dsa}
X.~Ning, T.~Zhao, W.~Li, P.~Lei, Y.~Wang, and H.~Yang, ``Dsa: More efficient
  budgeted pruning via differentiable sparsity allocation,'' in
  \emph{ECCV}.\hskip 1em plus 0.5em minus 0.4em\relax Springer, 2020, pp.
  592--607.

\bibitem{DBLP:pfp}
L.~Liebenwein, C.~Baykal, H.~Lang, D.~Feldman, and D.~Rus, ``Provable filter
  pruning for efficient neural networks,'' in \emph{ICLR}, 2020.

\bibitem{DBLP:drlbased}
J.~Chen, S.~Chen, and S.~J. Pan, ``Storage efficient and dynamic flexible
  runtime channel pruning via deep reinforcement learning,'' in \emph{NeurIPS},
  2020.

\bibitem{gao2020dmc}
S.~Gao, F.~Huang, J.~Pei, and H.~Huang, ``Discrete model compression with
  resource constraint for deep neural networks,'' in \emph{CVPR}, 2020, pp.
  1899--1908.

\bibitem{chin2020legr}
T.-W. Chin, R.~Ding, C.~Zhang, and D.~Marculescu, ``Towards efficient model
  compression via learned global ranking,'' in \emph{CVPR}, 2020, pp.
  1518--1528.

\bibitem{gao2021nppm}
S.~Gao, F.~Huang, W.~Cai, and H.~Huang, ``Network pruning via performance
  maximization,'' in \emph{CVPR}, 2021, pp. 9270--9280.

\bibitem{DBLP:CGNet}
W.~Hua, Y.~Zhou, C.~D. Sa, Z.~Zhang, and G.~E. Suh, ``Channel gating neural
  networks,'' in \emph{NeurIPS}, 2019, pp. 1884--1894.

\bibitem{li2022ogcnet}
G.~Li, M.~Zhang, J.~Zhang, and Q.~Zhang, ``Ogcnet: Overlapped group convolution
  for deep convolutional neural networks,'' \emph{Knowledge-Based Systems},
  vol. 253, p. 109571, 2022.

\bibitem{ren2015fasterrcnn}
S.~Ren, K.~He, R.~Girshick, and J.~Sun, ``Faster r-cnn: Towards real-time
  object detection with region proposal networks,'' \emph{NeurIPS}, vol.~28,
  pp. 91--99, 2015.

\bibitem{lin2017retinanet}
T.-Y. Lin, P.~Goyal, R.~Girshick, K.~He, and P.~Doll{\'a}r, ``Focal loss for
  dense object detection,'' in \emph{ICCV}, 2017, pp. 2980--2988.

\bibitem{wu2019detectron2}
Y.~Wu, A.~Kirillov, F.~Massa, W.-Y. Lo, and R.~Girshick, ``Detectron2,''
  \url{https://github.com/facebookresearch/detectron2}, 2019.

\bibitem{elsken2019nas}
T.~Elsken, J.~H. Metzen, and F.~Hutter, ``Neural architecture search: A
  survey,'' \emph{The Journal of Machine Learning Research}, vol.~20, no.~1,
  pp. 1997--2017, 2019.

\bibitem{lin2017rnp}
J.~Lin, Y.~Rao, J.~Lu, and J.~Zhou, ``Runtime neural pruning,'' in
  \emph{NeurIPS}, 2017, pp. 2178--2188.

\bibitem{akhlaghi2018snapea}
V.~Akhlaghi, A.~Yazdanbakhsh, K.~Samadi, R.~K. Gupta, and H.~Esmaeilzadeh,
  ``Snapea: Predictive early activation for reducing computation in deep
  convolutional neural networks,'' in \emph{ACM/IEEE 45th Annual International
  Symposium on Computer Architecture (ISCA)}.\hskip 1em plus 0.5em minus
  0.4em\relax IEEE, 2018, pp. 662--673.

\end{thebibliography}

\end{document}